\documentclass[10pt,twocolumn,letterpaper]{article}
\usepackage{twemojis}

\usepackage[
  letterpaper,
  top=0.72in,
  bottom=0.78in,
  left=0.66in,
  right=0.66in,
  columnsep=0.24in
]{geometry}
\usepackage[T1]{fontenc}
\usepackage[utf8]{inputenc}
\usepackage{times}
\usepackage{microtype}
\usepackage{graphicx}
\usepackage{booktabs}
\usepackage{multirow}
\usepackage{amsmath}
\usepackage{amssymb}
\usepackage{gensymb}
\usepackage{array}
\usepackage{ragged2e}
\usepackage{placeins}
\usepackage{abstract}
\usepackage[font=small,labelfont=bf]{caption}
\usepackage[numbers,sort&compress]{natbib}
\usepackage{xcolor}
\usepackage[
  breaklinks=true,
  colorlinks=true,
  linkcolor=blue,
  citecolor=blue,
  urlcolor=blue
]{hyperref}
\usepackage{xurl}  

\begin{document}


\title{%
  \texorpdfstring{%
    \twemoji[height=1.05em]{thinking face}\enspace
  }{Thinking face: }%
  Now We Know?\\
  A Systematic Comparison of TerraMind and THOR%
}
\author{
    Frederick~Schindlegger$^{1,2*}$,
    Kenzo~Bounegta$^{2,3*}$,
    Eva~Gmelich~Meijling$^{2}$,
    Johannes~Jakubik$^{4}$,\\
    Arnt-B\o rre~Salberg$^{5}$,
    Theodor~Forgaard$^{5}$,
    Nicolas~Longepe$^{2}$,
    and~Valerio~Marsocci$^{2}$\\[0.8em]
    {\small $^{1}$University of Münster, Münster, Germany}\\
    {\small $^{2}$ESA $\Phi$-lab, ESRIN, Frascati, Italy}\\
    {\small $^{3}$Rubicon, Paris, France}\\
    {\small $^{4}$IBM Research, Zurich, Switzerland}\\
    {\small $^{5}$Norwegian Computing Center, Oslo, Norway}\\[0.5em]
    {\small $^{*}$Equal contribution.}
}
\date{}
\twocolumn[
\maketitle
\begin{onecolabstract}
Benchmarks for Geospatial Foundation Models (GFMs) increasingly rank
models by aggregate score, but such rankings obscure \emph{why} models
differ: how much of the gap is architecture, how much is decoder
capacity, and how much is a use-case-specific artefact? This study addresses that gap through
a controlled comparison of two GFMs developed under European Space
Agency's $\Phi$-lab with contrasting design
philosophies: THOR, which introduces a compute-adaptive architecture
supporting variable patch sizes and unifies Sentinel-1, -2, and -3 data
at their native resolutions; and TerraMind, a multimodal generative GFM
pretrained with a dual-scale token/pixel objective that enables
any-to-any cross-modal generation (Thinking-in-Modalities) to
infer missing sensors at inference time. Rather than reporting a single
leaderboard, we investigate the axes along which the two architectures
actually differ -- patch size, decoder complexity, finetuning regime, input modality, and model scale -- across ten use cases
spanning segmentation and regression in diverse domains, including climate disaster response, methane
leak detection, snow monitoring, or sea ice mapping.
We find that \textit{architectural
design choices -- patch size and decoder type in particular -- explain
more performance variance than model identity itself}, that the two
models embody complementary investment strategies (pretraining-time
scale for TerraMind versus inference-time tokenisation for THOR), and
that correctly interpreting results
requires dataset-level characterisation. The
resulting picture is not a single winner but a set of hypotheses and a
diagnostic ablation methodology that we expect to generalise to future
GFMs beyond THOR and TerraMind.
\end{onecolabstract}

\vspace{0.35em}
\begin{center}
\end{center}
\vspace{0.8em}
]


\section{Introduction}

As Earth Observation (EO) data archives continue to expand by petabytes
each year, extracting meaningful information from such large,
heterogeneous datasets remains a major bottleneck. Machine Learning
offers powerful tools for EO analysis, and task-specific supervised
models can achieve high accuracy when little labelled data is available.
However, their performance often degrades when applied beyond their
training domain, which poses challenges in generalising across sensors, resolutions, and geographies.

Geospatial Foundation Models (GFMs) have emerged as a promising
solution \cite{11303008}. By pretraining on large, unlabeled, multisensor datasets,
they aim to produce general-purpose representations that transfer
effectively to a wide range of downstream tasks. Over the past three years, the number of GFMs has grown
substantially. Notable examples include the Prithvi
family~\cite{szwarcman2024prithvi}, Olmo-Earth~\cite{herzog2025olmoearthstablelatentimage},
and an expanding class of Earth-embedding models such as
AlphaEarth~\cite{brown2025alphaearth}. Despite this progress, gaps
remain, particularly in thorough cross-model comparison and
understanding how architectural design choices influence real-world
performance.

For EO practitioners, the ability of a model to generalise spatially and
temporally outside its training distribution is essential. Foundation
models must support diverse input modalities, maintain robustness under
operational conditions, and demonstrate value beyond curated benchmark
tasks.

To address these needs, the European Space Agency (ESA), through $\Phi$-lab, has funded two
complementary GFM initiatives.
THOR~\cite{forgaard2026thor}, introduces a Vision Transformer (ViT) architecture
that supports variable patch sizes at inference time, allowing dynamic
trade-offs between computational cost and feature resolution without
retraining. It is the first GFM to unify Copernicus Sentinel-1 (SAR),
Sentinel-2 (multispectral), and Sentinel-3 (OLCI/SLSTR) data at their
native resolutions (10\,m--1\,km) within a single model, pretrained on a
new 22\,TB multi-sensor dataset.
TerraMind~\cite{jakubik2025terramind}, is a multimodal generative model pretrained on 500 billion tokens
from the TerraMesh dataset~\cite{blumenstiel2025terramesh}, which includes Sentinel-1 SAR imagery, Sentinel-2 optical imagery, land use/land cover (LULC) maps, normalized difference vegetation index (NDVI) maps, digital elevation models (DEM), geographic coordinates, and natural language captions. Its dual-scale transformer encoder--decoder captures both
token-level context and pixel-level spatial detail, and introduces
Thinking-in-Modalities (TiM), a mechanism enabling the generation of
missing modalities as intermediate reasoning steps during fine-tuning and
inference.

The two models show contrasting design philosophies. THOR emphasises compute
adaptability and heterogeneous sensor integration, aimed to enable
data-limited settings and lightweight decoder usage.
TerraMind prioritises rich cross-modal representations and generative
capabilities, enabling interpolation of
missing input modalities. This paper provides a systematic evaluation of these
trade-offs.

Existing GFM benchmarks are well suited to answering \emph{which model
ranks highest}, but poorly suited to answering \emph{why}. A recent study auditing 152 GFM papers \cite{corley2026knowsstateartgeospatial} claims that no one knows the current state of the art in GFMs, finding cross-paper disagreements exceeding ten points for the same model, benchmark, and protocol, alongside widespread confounding of architecture and pretraining-data changes. Aggregate leaderboards average performance across curated task collections,
which conflates the contribution of the pretrained backbone with that
of the decoder, the fine-tuning regime, and the spatial tokenisation
scheme, and offers little guidance on whether an observed gap reflects
a genuine architectural advantage or a dataset-specific artefact. THOR
and TerraMind are particularly well suited to closing this gap because
they instantiate two deliberately contrasting design philosophies under
a shared funding programme and a shared evaluation framework
(TerraTorch~\cite{gomes2025terratorch}). Treating this pair as
a controlled natural experiment makes it possible to attribute performance differences to
specific, nameable design axes, and to ask which of these axes are
likely to matter for GFMs more generally.

The contributions of this paper are fourfold.
\begin{enumerate}
    \item We run a large controlled ablation across two architecturally contrasting GFMs and ten EO use cases spanning diverse domains.
    \item We introduce a diagnostic methodology that isolates the marginal contribution of architectural deployment choices.
    \item We show that correctly interpreting ablation results requires use-case-level characterisation, and that such characterisation is a necessary component of rigorous GFM benchmarking rather than a preprocessing detail.\
    \item We distil the resulting patterns into hypotheses that we expect to generalise to ViT-based GFMs beyond THOR and TerraMind, alongside practical deployment guidance for EO practitioners choosing between GFM architectures.
\end{enumerate}


\begin{table*}[t]
\centering
\caption{Architectural and Operational Comparison of THOR and TerraMind (GSD stands for Ground Sample Distance)}
\label{tab:model_comparison}
\renewcommand{\arraystretch}{1.3} 
\footnotesize
\resizebox{0.85\textwidth}{!}{%
\begin{tabular}{lll}
\toprule
\textbf{Design Axis} & \textbf{THOR} & \textbf{TerraMind} \\
\midrule
Consortium / Origin & FM4CS (NR, UiT, ESA $\Phi$-lab) & FAST-EO (DLR, Jülich, IBM, KP Labs, ESA $\Phi$-lab) \\
Core Philosophy & Compute-adaptive native GSD fusion & Generative cross-modal representation \\
Primary Model Type & Vision Transformer (FlexiViT) & Transformer Encoder-Decoder \\
Sensors \& Modalities & Sentinel-1, -2, -3 (OLCI/SLSTR) & Sentinel-1, -2, LULC, DEM, Text \\
Spatial Granularity & Variable patch size (ps $ \in \{4, 8, 16, 32\}$) & Fixed patch size (ps $= 16$) \\
GSD Handling & Native $10\,\text{m}$ to $1\,\text{km}$ (GSD-aware) & Nominal $10\,\text{m}$--$20\,\text{m}$ ($224^2$ minimum) \\
Modality Fusion & Early channel-wise concatenation & Token-level pooling with projection \\
Position Encoding & 2D-ALiBi (FlexAttention-ready) & Standard 2D positional embedding \\
Distinctive Capability & Dynamic inference-time trade-offs & Thinking-in-Modalities (TiM) \\
\bottomrule
\end{tabular}%
}
\end{table*}
\section{Related Work}
\subsection{Geospatial Foundation Models}
\label{sec:related_gfms}

The foundation model concept, formalised in~\cite{bommasani2021opportunities}, was adapted to EO around 2022 through unimodal self-supervised learning (SSL) GFMs. Two dominant computer vision SSL paradigms emerged: contrastive learning~\cite{he2020momentum,caron2021emerging} to align augmented views, and Masked Image Modelling (MIM) -- including Masked Autoencoders (MAE)~\cite{he2022masked} -- to reconstruct masked patches. Both paradigms typically build on ViT backbones. EO-specific SSL extended these concepts to multi-spectral/temporal dimensions (e.g., SatMAE~\cite{cong2022satmae}) and varying spatial resolutions (e.g., Scale-MAE~\cite{reed2023scale}). These early approaches, however, remained constrained to single-sensor or fixed-spectral configurations.

To address these limitations, a second wave of multimodal GFMs emerged. CROMA~\cite{fuller2023croma} fused Sentinel-1 and Sentinel-2 data using contrastive radar-optical objectives. DOFA~\cite{xiong2024neural} introduced wavelength-conditioned layers for arbitrary spectral projections, processing multispectral, hyperspectral, or SAR inputs within a single model. MMEarth~\cite{nedungadi2024mmearth} aligned co-located multi-modal observations, proving that pretraining data diversity is as critical as architectural scaling. AnySat~\cite{astruc2025anysat} enabled complete flexibility across arbitrary spatial and spectral GSDs, while HyperSIGMA~\cite{wang2025hypersigma} introduced specialized spatial-spectral fusion modules for contiguous hyperspectral bands. Concurrently, the Prithvi family~\cite{jakubik2023prithvi} scaled MAE to HLS time series, later expanded by Prithvi-EO-2.0~\cite{szwarcman2024prithvi} across broader sensors and geographies.

\textbf{Compute-adaptive tokenisation.} Rather than spectral adaptability, a distinct lineage targets spatial granularity. FlexiViT~\cite{beyer2023flexivit} proved that a single ViT backbone can operate across variable patch sizes at inference, trading compute for local resolution post-hoc. OlmoEarth~\cite{herzog2025olmoearthstablelatentimage} implemented compute-adaptive processing within an open, auditable workflow. THOR~\cite{forgaard2026thor} represents the most complete EO instantiation of this lineage, combining FlexiViT-style patching with GSD-aware positional encoding to natively unify Sentinel-1, -2, and -3 at their respective native resolutions.

\textbf{Generative GFMs.} Alternatively, generative pretraining has progressed from RGB-focused diffusion models like DiffusionSat~\cite{khanna2023diffusionsat} to multimodal generative systems~\cite{espinosa2026cop}. Features from generative denoising have been shown to rival standard SSL encoders for downstream dense prediction~\cite{jia2025can}. TerraMind~\cite{jakubik2025terramind} is a premier generative GFM in this space, trained on 500 billion tokens over eight modalities with a dual-scale token/pixel objective. It leverages TiM to synthesize missing inputs or auxiliary layers dynamically during inference.

\textbf{Vision-Language Models.} Vision-language models like RemoteCLIP~\cite{liu2024remoteclip} and GeoRSCLIP~\cite{zhang2024rs5m} aim to integrate natural language and metadata into a unified multi-modal space. 

\textbf{Embedding Models.} Concurrently, decoupled embedding architectures like AlphaEarth~\cite{brown2025alphaearth} and TESSERA~\cite{feng2025tessera} bypass downstream task-specific training entirely by providing precomputed pixel-level spatial-temporal embeddings directly on EO platforms.

\subsection{Benchmarking GFMs}
\label{sec:related_bench}

The rapid growth of GFMs necessitates rigorous benchmarking. GEO-Bench~\cite{lacoste2023geo} initiated multi-task evaluations against task-specific baselines, while PANGAEA~\cite{marsocci2024pangaea} established standardized protocols using fixed decoders and hyperparameters to isolate representation quality. Specialized benchmarks have since targeted distinct capability dimensions: MMEarth-Bench~\cite{gordon2026mmearth} probes cross-modal transfer on paired datasets, and GEO-Bench-2 \cite{simumba2025geo} evaluates fine-grained capabilities across nineteen datasets. Domain-specific suites have also emerged, such as FoMo-Bench~\cite{bountos2025fomo} for forest monitoring and Cryo-Bench~\cite{kaushik2026cryo} for cryospheric mapping, showing that domain shifts can heavily alter GFM rankings. Today, TerraTorch~\cite{gomes2025terratorch} provides a unified, open framework standardizing GFM fine-tuning and evaluation, serving as the core environment for this study.

\vspace{0.2cm}
These efforts have been instrumental in establishing shared evaluation
norms and providing aggregate rankings across models, but they are
designed to answer a \emph{ranking} question -- which model achieves the
highest average score across a curated task suite -- and not an
\emph{attribution} question: given an observed performance gap between 
models, how much is due to the backbone's pretraining objective, how
much to the decoder paired with it, how much to the spatial tokenisation
scheme, and how much to idiosyncrasies of the specific datasets used for
evaluation? Because aggregate rankings fix these factors implicitly
(typically to a single decoder and a single input resolution per model,
as in PANGAEA's protocol), they cannot disentangle them, and the
reported ranking is consequently protocol-dependent in ways that are
rarely made explicit. This matters in practice: Cryo-Bench and
MMEarth-Bench already show that GFM rankings shift substantially when
moving from general-purpose to domain-specific or cross-modal
evaluation -- evidence that what aggregate rankings capture is not a
stable, transferable property of the model, but an interaction between
model, decoder, task, and dataset.

Answering \emph{what drives performance} therefore requires a different
evaluation design from answering \emph{which model wins}: one that
treats architectural and training-configuration choices as independent
variables to be studied, and
that couples this ablation with dataset-level scrutiny. Motivated by this
requirement, we run a large number of controlled experiments across two
models with contrasting architectures, varying patch size, decoder
complexity, backbone adaptation regime, and input modality
systematically, using THOR and TerraMind as the controlled pair through
which the attribution question is posed.


\section{Use Cases}
\label{sec:usecases}
This study spans ten use cases adapted from the two ESA consortia -- FAST-EO (TerraMind) and FM4CS (THOR) -- with five from each consortium, covering semantic segmentation, regression, object detection, and change detection. Table~\ref{tab:usecases} provides an overview of the benchmarked datasets and their diversity in sensing modalities, spatial resolutions, input configurations, and prediction objectives. More details about the datasets are in Appendix \ref{app:usecases}.

Owing to their differing objectives, the use cases contributed by the two consortia are complementary. The FAST-EO use cases include well-known classical benchmark datasets, feature multimodal setups that combine remote sensing products with auxiliary data, and span a global and diverse range of domains, though all five are variants of semantic segmentation. The FM4CS use cases, in contrast, target novel and less well-studied problems, with a particular focus on the arctic domain and a stronger emphasis on SAR data; they also incorporate Sentinel-3 (S3) data, a modality that remains less established for GFMs, and cover a more diverse range of task types, including regression and segmentation. Together, these two sets of use cases provide broader coverage of domains, modalities, task types, and task difficulty than either consortium would offer individually.

\begin{table*}[t]
\centering
\caption{FAST-EO and FM4CS use cases overview.}
\label{tab:usecases}
\resizebox{\textwidth}{!}{%
\begin{tabular}{c|lllllll}
\toprule
& \textbf{Dataset} & \textbf{Sensor(s)} & \textbf{GSD} & \textbf{Patch}
  & \textbf{Task} & \textbf{Target} & \textbf{Metric} \\
\midrule
\multirow{5}{*}{\rotatebox[origin=c]{90}{FAST-EO}}
& Sen1Floods11      & S1 + S2                  & 10\,m  & $512^2$  & Segmentation   & Flood / no flood       & mIoU \\
& HLS Burn Scars    & S2 (HLS)                 & 30\,m  & $512^2$  & Segmentation   & Burned / unburned      & mIoU \\
& CocoaMining       & S1 + S2 + DEM            & 10\,m  & $128^2$  & Segmentation   & Mines / cocoa / background & mIoU \\
& Methane Leaks     & Hyperspectral (airborne) &  --     & $512^2$  & Classification & Methane plume          & OA \\
& HYPERVIEW         & Hyperspectral (airborne) &  --     &  --      & Regression     & K, Mg, P$_2$O$_5$, pH & RMSE \\
\midrule
\multirow{5}{*}{\rotatebox[origin=c]{90}{FM4CS}}
& Flood Zone        & S1                       & 10\,m  & $128^2$  & Segmentation   & Flood / no flood       & F1 \\
& Iceberg Detection & S1                       & 10\,m  & $256^2$  & Segmentation   & Iceberg / background   & F1 \\
& Sea Ice           & S1                       & 250\,m & $224^2$  & Segmentation   & 3 classes              & mIoU \\
& Snow Monitoring   & S3 SLSTR                 & 500\,m & $144^2$  & Regression     & Snow cover             & RMSE \\
& Mires/Wetlands Mapping  & S2                       & 10\,m  & $288^2$  & Segmentation   & 6 classes              & mIoU \\
\bottomrule
\end{tabular}%
}
\end{table*}


\section{Methodology}
\label{sec:methodology}

\subsection{Evaluation Philosophy}
\label{sec:eval_philosophy}

We adopt a controlled ablation study following the standardised
evaluation philosophy of PANGAEA~\cite{marsocci2024pangaea}: all
hyperparameters not intrinsic to a model's architecture are fixed, while
each model may exploit its own structural specificities e.g. variable patch
sizes for THOR. All experiments
are conducted within TerraTorch~\cite{gomes2025terratorch}.

Crucially, we control for model capacity rather than architectural configuration: we do not degrade the operating conditions of one model to
match the hardware limitations of another. Each model is allowed to run
at its best achievable configuration within the constraints of a single
RTX~A6000 GPU (46\,GB VRAM). We acknowledge that this principle has
limits: matching models by capacity does not necessarily yield fair comparisons, and differences
in pretraining data volume, modality coverage, and optimisation recipe
mean that any comparison reflects the full model pipeline, not the
backbone alone. The search space is also inherently
asymmetric: THOR's variable patch size introduces four additional
configurations per experiment cell (ps4/8/16/32), while TerraMind
operates at a fixed patch size of~16 with no equivalent axis of
variation. As a result, THOR has a larger configuration space and more
opportunities to find a well-tuned operating point.

Owing to computational constraints and use-case-specific prerequisites, not all ablation studies are performed across all use cases. Certain analyses are selectively omitted where they are either computationally prohibitive or not applicable to the characteristics of a given task. 

\subsection{Decoder Design}
\label{sec:decoders}

We compare two decoders of contrasting capacity to disentangle backbone
representation quality from decoder sophistication:

\begin{enumerate}
    \item \textbf{Linear.} A single projection from the backbone's final
    layer to the segmentation logits (256 channels), followed by learned
    upsampling to the input resolution. This minimal decoder isolates the
    backbone's representation quality: any performance gain over a random
    encoder is directly attributable to the pre-trained features. Note
    that the number of input tokens also influences this decoder: a
    higher token count (finer patch size or larger input resolution)
    provides the linear projection with spatially denser features,
    independently of the backbone's representational quality.

    \item \textbf{UperNet.} The UPerNet decoder~\cite{xiao2018unified}
    with Pyramid Pooling Module (pool scales $\{1,2,3,6\}$) and a
    lateral-connection Feature Pyramid Network. It receives activations
    from four intermediate backbone layers (indices $\{2,5,8,11\}$ for a
    12-block ViT-Tiny or ViT-Base), reshaped into a multi-scale pyramid
    via a learned upsampling neck. This configuration is consistent with
    the decoder used in PANGAEA~\cite{marsocci2024pangaea} and in the
    TerraMind paper~\cite{jakubik2025terramind}.
\end{enumerate}

Both decoders use 256 hidden channels and dropout~0.2. For the FM4CS
use cases of flood change detection and iceberg object detection, we evaluate lightweight decoders: (i)~a \textbf{PixelShuffle Linear}~\cite{shi2016realtime} decoder applying sub-pixel upsampling for efficient
high-resolution reconstruction; and (ii)~a \textbf{Multi-Layer
Perceptron (MLP)}~\cite{xie2021segformer} decoder mapping backbone features directly to logits.

\subsection{Spatial Handling and Ground Sampling Distance}
\label{sec:spatial_handling}
Each model encodes assumptions about spatial resolution.
THOR~\cite{forgaard2026thor} communicates the physical ground extent of
each input tile via a \texttt{ground\_covers} parameter (in metres),
conditioning its ALiBi-based positional encoding on the true ground
sampling distance (GSD). TerraMind~\cite{jakubik2025terramind}
tokenizer models are pre-trained on input images ranging from
$224{\times}224$ to $256{\times}256$~px, using a ViT encoder with patch
size~16 and a patched UNet decoder with patch size~4.
These constraints lead to different spatial strategies depending on
image size and native GSD:

\textbf{Large images, exceeding TerraMind's native pretraining range}
(Sen1Floods11, HLS Burn Scars, Mires/Wetlands).
Both models receive $288 \times 288$~px crops during training. For THOR,
the \texttt{ground\_covers} parameter is set to the true physical extent
of the crop: $2\,880$\,m for Sen1Floods11 ($288 \times 10$\,m) and
$8\,640$\,m for HLS Burn Scars ($288 \times 30$\,m). For Mires/Wetlands, we
use tiled training and inference with overlapping windows
(\texttt{h\_crop}~=~\texttt{w\_crop}~=~288, \texttt{h\_stride}~=~\texttt{w\_stride}~=~256);
as with the other datasets in this group, THOR's \texttt{ground\_covers}
is set to match the physical extent of the $288\times288$ crop at
Mires/Wetlands' native GSD. For spatial fairness, TerraMind receives the same
$288\times288$~crops rather than its nominal $224\times224$
pretraining size. Although this exceeds the pretraining input range,
it increases the token count from 196 to 324 ($288/16 = 18$ tokens per
side), which improves performance on tiled inference tasks by providing
the self-attention layers with more spatial context. At validation and
test time, both models evaluate full $512\times512$ images via
sliding-window (tiled) inference with overlapping strides.

\textbf{Large images, within TerraMind's native pretraining range}
(Flood Zone Change Detection and Iceberg Detection at 10\,m GSD).
Both models receive $256 \times 256$~px crops during training and
evaluation, which lies within TerraMind's native pretraining range and
therefore requires no resizing. For THOR, \texttt{ground\_covers} is
set to $2\,560$\,m ($256 \times 10$\,m GSD) for both datasets.

\textbf{Sea Ice} (10\,m GSD). Images are processed at $224\times224$~px,
which corresponds exactly to TerraMind's minimum native pretraining
resolution and therefore requires no resizing or upsampling. For THOR,
\texttt{ground\_covers} is set to $2\,240$\,m ($224 \times 10$\,m GSD),
consistently reflecting the true physical extent of the input.

\textbf{Small images} ($< 224\times224$~px; CocoaMining at $128\times128$, Snow
at $144\times144$). For THOR, we preserve the native $128\times128$
resolution for CocoaMining and set \texttt{ground\_covers}~$= 1\,280$\,m
($= 128 \times 10$\,m GSD), faithfully reflecting the physical ground
extent. For TerraMind, $128\times128$ is below the minimum
pretraining resolution; images are therefore resized to $224 \times 224$,
the smallest pretraining input size. The snow use case operates on
fixed input size $144\times144$~px, corresponding to its native
Sentinel-3 SLSTR resolution. No resizing or upsampling is applied for
either THOR or TerraMind.

We note that resizing THOR's CocoaMining inputs to $224\times224$ (with
appropriately adjusted \texttt{ground\_covers}) improves performance,
likely because the increased token count ($196$ vs.\ $64$ at $128\times128$
with ps16) provides richer spatial context. In either case, THOR's
\texttt{ground\_covers} is always adjusted to match the physical extent
of the actual input, ensuring GSD-aware 2D-ALiBi positional encoding remains
consistent with pretraining assumptions.

\textbf{Note on spatial fairness.} The different spatial handling
strategies described above mean that the two models do not always receive
geometrically identical inputs, particularly on CocoaMining where
TerraMind is resized to $224\times224$ while THOR operates at $128\times128$.
Depending on the dataset and the degree of upsampling, this spatial
asymmetry may advantage or disadvantage one model. Results on such
datasets should therefore be interpreted with this caveat in mind.

\subsection{Model-Specific Considerations}
\label{sec:model_specific}

\textbf{THOR -- variable patch size.} THOR's FlexiViT
backbone~\cite{beyer2023flexivit} accepts a runtime patch size $ps \in \mathbb{N}^{+}$, with commonly used values typically satisfying $4 \leq ps \leq 32$, tiling each image into $n = (H/ps)^2$ tokens (where $H$ stand for height i.e. one of the two dimension of the image). Smaller
patches produce higher-resolution feature maps at the cost of
quadratically longer self-attention sequences, spanning a $64\times$
range in token count across the investigated patch sizes.

\textbf{TerraMind -- fixed resolution and Thinking-in-Modalities.}
TerraMind operates at a fixed patch size of~16. Its distinctive
capability is TiM, which generates synthetic
modality tokens at inference time -- either reconstructing an ablated
sensor input or producing a task-relevant auxiliary modality (e.g.\
LULC). The results about TiM are discussed in the Appendix \ref{sec:tim_results}.

\textbf{Multimodal fusion.} For datasets providing co-registered
Sentinel-1 and Sentinel-2 acquisitions (Sen1Floods11, CocoaMining), we
evaluate S1-only, S2-only, and S1+S2 configurations; for CocoaMining,
TerraMind additionally tests S1+S2+DEM. THOR concatenates all bands
along the channel dimension before a shared patch projection; TerraMind
processes each modality through independent projection heads and fuses
representations at the token level via mean pooling.

\textbf{Task-specific baseline (UNet).} To contextualise GFM performance
on the FAST-EO use cases, we include a standard UNet encoder-decoder
trained from scratch (no pretraining) on Sen1Floods11, HLS Burn Scars,
and CocoaMining. The UNet baseline is not a primary focus of this
comparison. UNet serves to quantify at what annotation budget GFM
pretraining provides a practical advantage, particularly in the
low-data regime. Full-dataset UNet results are reported in
Appendix~\ref{app:unet_results}.

\subsection{Encoder Computational Cost}
\label{sec:flops}

Table~\ref{tab:flops} reports the encoder-only forward-pass cost for
each configuration on a single $288 \times 288$~px input, in
GMACs. THOR is profiled with
\texttt{torch.utils.flop\_counter}; TerraMind with
\texttt{fvcore}~\cite{fvcore}. All values exclude the decoder.

\begin{table}[t]
\centering
\caption{Encoder-only cost (GMACs) on a $288\times288$~px input (S2: 10~bands,
S1: 2~bands), profiled empirically on Sen1Floods11.
Decoder cost excluded.  Token counts are empirical (include all modality
tokens); the theoretical $(H/p)^2$ formula underestimates the actual
sequence length.
For CocoaMining, THOR operates at $128\times128$ (empirical token counts:
$2304$, $576$, $144$, $36$ at ps4/8/16/32) and TerraMind operates at
$224\times224$ ($196$~tokens, ${\sim}50$~GMACs encoder).}
\label{tab:flops}
\begin{tabular}{lrrr}
\toprule
\textbf{Configuration} & \textbf{Tokens} & \textbf{GMACs}
  & \textbf{Params\,(M)} \\
\midrule
THOR ps32      &    178     &    18  & 94.1 \\
THOR ps16      &    729     &    62  & 94.1 \\
THOR ps8       &  2\,916    &   248  & 94.1 \\
THOR ps4       & 11\,664    &   991  & 94.1 \\
\midrule
TerraMind ps16 &    324     &    56  & 87.9 \\
\bottomrule
\end{tabular}
\end{table}

At ps16 and $288\times288$, both encoders are nearly iso-cost -- 62 vs.\
56~GMACs -- despite different architectures and parameter counts.
Halving the patch size roughly quadruples the encoder GMACs, consistent
with the quadratic scaling of self-attention.

\textbf{Caveats.} (i)~The linear decoder adds negligible overhead;
UperNet's FPN + PPM is more substantial but remains modest relative to
the encoder at small patch sizes. (ii)~For CocoaMining, TerraMind
operates at $224\times224$ (196~tokens, ${\sim}50$~GMACs encoder,
${\sim}61$~GMACs with UperNet) while THOR operates at $128\times128$ (token
count varies by patch size), so the effective per-sample costs differ
from the $288\times288$ reference in Table~\ref{tab:flops}.

\section{Experimental Setup}
\label{sec:setup}

\subsection{Training Configuration}
\label{sec:training_config}

All experiments run on a single NVIDIA RTX~A6000 GPU (46\,GB VRAM) with
16-bit mixed precision (32 for snow mapping as an exception).  The experiment configurations are available at
\url{https://github.com/KenzoBou/Terramind-vs-Thor-ESA-PhiLab}. All runs use AdamW~\cite{loshchilov2019adamw}
with initial learning rate $2 \times 10^{-5}$, weight decay~$0.05$. The
learning rate is reduced on plateau (\texttt{ReduceLROnPlateau},
factor~$0.5$, patience~10~epochs, floor~$10^{-6}$); early stopping
monitors validation loss with patience~30; training runs for at most
200~epochs. The best checkpoint is selected as the epoch with the lowest
validation loss; the test set is evaluated only once using this
checkpoint, and is never used during hyperparameter selection.
In the frozen-backbone regime, the learning rate is
multiplied by~5 to accelerate decoder convergence, following
PANGAEA~\cite{marsocci2024pangaea}. All experiments use seed~$=0$. 

\textbf{Note on single-seed evaluation.} All experiments are conducted
with a single random seed. Given the relatively small performance
differences between models on several datasets, single-seed results
should be interpreted with caution: observed gaps of less than
${\sim}1$~pp may not be reproducible across seeds. 

Data augmentation for the FAST-EO use cases consists of D4 transforms
(random flips and $90\degree$ rotations) applied during training. The
FM4CS use cases use a content-aware sampling strategy via
\texttt{albumentations.CropNonEmptyMaskIfExists} to preferentially
select informative regions containing target annotations.

\subsection{Ablation Grid}
\label{sec:ablation_grid}

Tables~\ref{tab:ablation_grid_all}
summarises the axes of variation for the investigated FAST-EO and FM4CS
datasets, respectively.
For the FAST-EO datasets, the dominant axis of variation is THOR's patch size, which spans a $64\times$ range in token count from ps32 to ps4. Decoder type (Linear vs.\ UperNet) is the second most influential axis, particularly at coarse patch sizes where backbone features are spatially under-resolved. Backbone freeze regime has the smallest marginal effect for TerraMind, but matters more for THOR at intermediate patch sizes. For the FM4CS datasets, the axes follow the same structure.

\begin{table*}[t]
\centering
\caption{Ablation grid across all datasets. TerraMind uses a fixed
patch size of~16. `` -- '' = not applicable.}
\label{tab:ablation_grid_all}
\resizebox{\textwidth}{!}{%
\begin{tabular}{lllllll}
\toprule
\textbf{Dataset} & \textbf{Decoder} & \textbf{Backbone}
  & \textbf{THOR ps} & \textbf{THOR mod.} & \textbf{TM mod.}
  & \textbf{Data scarcity} \\
\midrule
Sen1Floods11
  & Linear, UperNet & Frozen, FT & 4, 8, 16, 32
  & S1, S2, S1{+}S2 & S1, S2, S1{+}S2
  & 5\%, 10\%, 25\%, 50\% \\
HLS Burn Scars
  & Linear, UperNet & Frozen, FT & 4, 8, 16, 32
  & S2 & S2
  & 5\%, 10\%, 25\%, 50\% \\
CocoaMining
  & Linear, UperNet & Frozen, FT & 4, 8, 16, 32
  & S1, S2, S1{+}S2 & S1, S2, S1{+}S2, +DEM
  &  -- \\
Iceberg Detection & PixelShuffle, MLP & Frozen & 4, 8, 16, 32& S1 & S1 &  -- \\
Snow Monitoring   & Linear, UperNet & Frozen, FT & 4, 8, 16, 32 & S3 & S3(Mapped to S2) &  -- \\
Mires/Wetlands Mapping  & Linear, UperNet & Frozen & 4, 8, 16, 32 & S2 & S2 &  -- \\
Flood Zone Detection  & PixelShuffle, MLP & Frozen & 4, 8, 16, 32 & S1 & S1 &  -- \\
Sea Ice Mapping  & Linear, UperNet & Frozen, FT & 4, 8, 16, 32 & S1 & S1 &  -- \\
\bottomrule
\end{tabular}%
}
\end{table*}
\section{Experimental Results}
\label{sec:results}

We turn to the central empirical question of this study: \textit{under what
conditions does each model excel, and what are the dominant factors
governing downstream performance?} Rather than presenting results
model-by-model, we organise the analysis around a series of research
questions. Each subsection isolates one factor: compute budget, spatial
tokenisation, decoder design, backbone adaptation, input modality,
generative reasoning, and model scale; while controlling for the others.
This structure mirrors the ablation grid of
Section~\ref{sec:ablation_grid}.

All experiments use a
single seed ($=0$); we discuss implications for statistical reliability
in Section~\ref{sec:robustness}.

\begin{table*}[t]
\centering
\caption{%
  Full benchmark results, frozen backbone (top) vs.\ full fine-tuning (bottom).
  Segmentation: macro-averaged mIoU ($\uparrow$).
  Regression: RMSE ($\downarrow$).  Detection: instance-wise F1 ($\uparrow$).
  `` -- '' = not applicable.
  \textbf{Bold} = best result per column within each setting.}
\label{tab:full_results_merged}
\resizebox{\textwidth}{!}{%
\begin{tabular}{llcccccccc}
\toprule
\textbf{Setting} & \textbf{Model}
  & \textbf{Sen1Fl.} & \textbf{Burn Scars} & \textbf{Cocoa}
  & \textbf{Iceberg Det.} & \textbf{Snow}
  & \textbf{Mires/Wetlands} & \textbf{Sea Ice} & \textbf{Flood Zone} \\
  & & mIoU & mIoU & mIoU & F1 & RMSE & mIoU & mIoU & F1 \\
\midrule
\multirow{4}{*}{Frozen}
  & TerraMind Base 
    & 0.904 & 0.828 & 0.813
    & 0.823 & 9.190& \textbf{0.646} & 0.504 & 0.328\\
  & TerraMind Tiny 
    & 0.897 & 0.813 & 0.799
    & 0.604& 9.517&  --    & 0.494& 0.334\\
  & THOR Base 
    & \textbf{0.907} & \textbf{0.837} & \textbf{0.820}
    & \textbf{0.857}& 6.782& 0.636& 0.642& \textbf{0.730}\\
  & THOR Tiny 
    & 0.902 & 0.830 & 0.814
    & 0.770& \textbf{6.465}&  --    & \textbf{0.825}& 0.717\\
\midrule
\multirow{4}{*}{Fine-tuned}
  & TerraMind Base
    & \textbf{0.910} & \textbf{0.891} & 0.817
    &  -- & 8.956&  -- & 0.668&  --  \\
  & TerraMind Tiny
    & 0.902 & 0.872 & 0.807
    &  --    & 9.196&  --    & 0.755&  --  \\
  & THOR Base
    & 0.904 & 0.865 & \textbf{0.821}
    &  -- & \textbf{5.922}&  --  & \textbf{0.873}&  --  \\
  & THOR Tiny
    & 0.893 & 0.846 & 0.817
    &  --    & 6.437&  --    & 0.804&  --  \\
\bottomrule
\end{tabular}%
}
\end{table*}

\begin{figure*}[t]
\centering
\includegraphics[width=\textwidth]{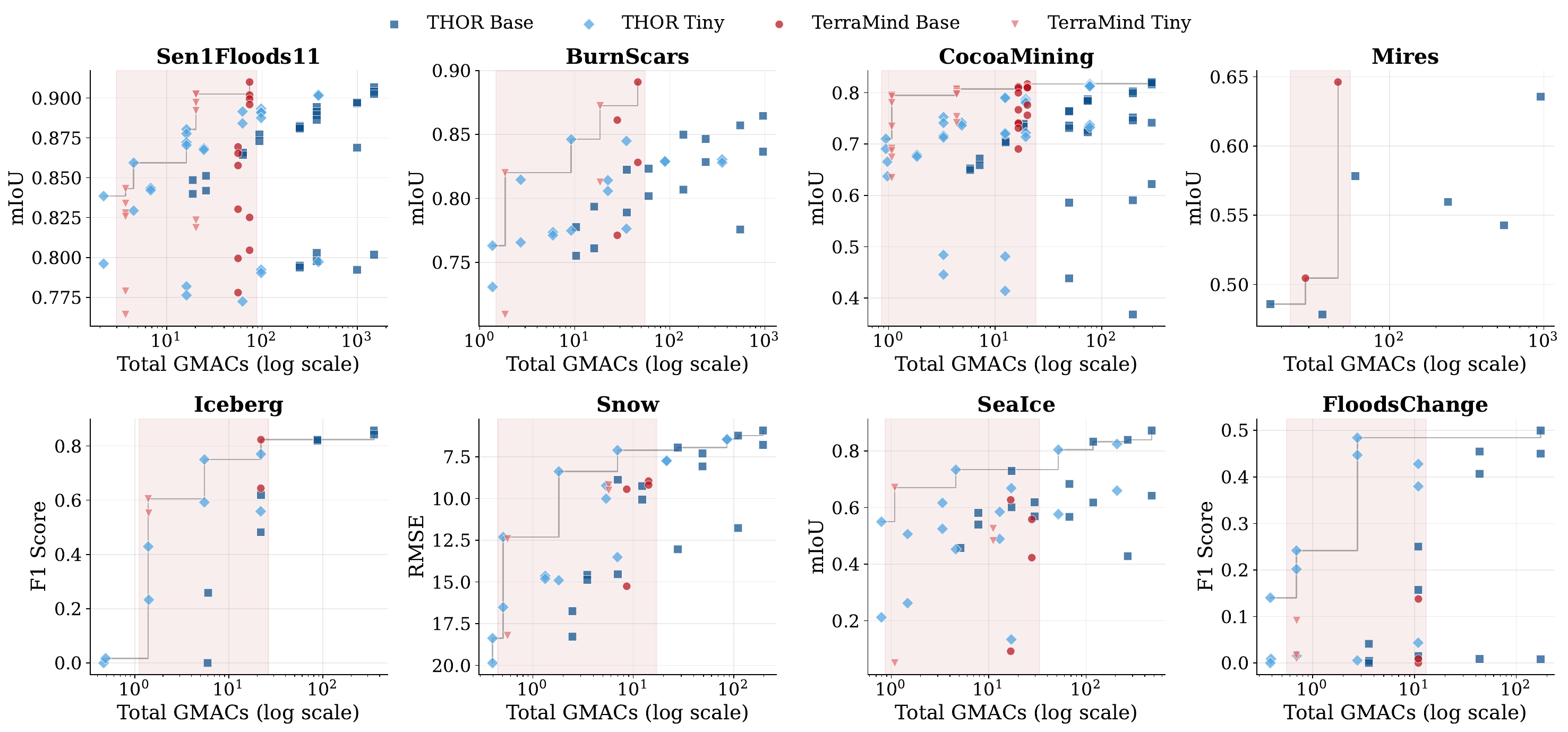}
\caption{%
  \textbf{Performance versus computational cost across benchmark datasets.}
  Panels show, for each dataset, task performance as a function of total
  pipeline compute (encoder~+~decoder GMACs, log scale; hyperspectral
  datasets excluded). Each marker represents one (model, decoder, patch
  size, freeze-state) configuration at the Base parameter scale; marker
  color denotes model and marker shape denotes configuration variant
  (see legend). Shaded band indicates the approximate GMACs range spanned
  by TerraMind configurations. Black step line traces the Pareto frontier
  (configurations not dominated by any other in performance--cost space).
}
\label{fig:pareto}
\end{figure*}

\begin{figure}[t]
\centering
\includegraphics[width=\columnwidth]{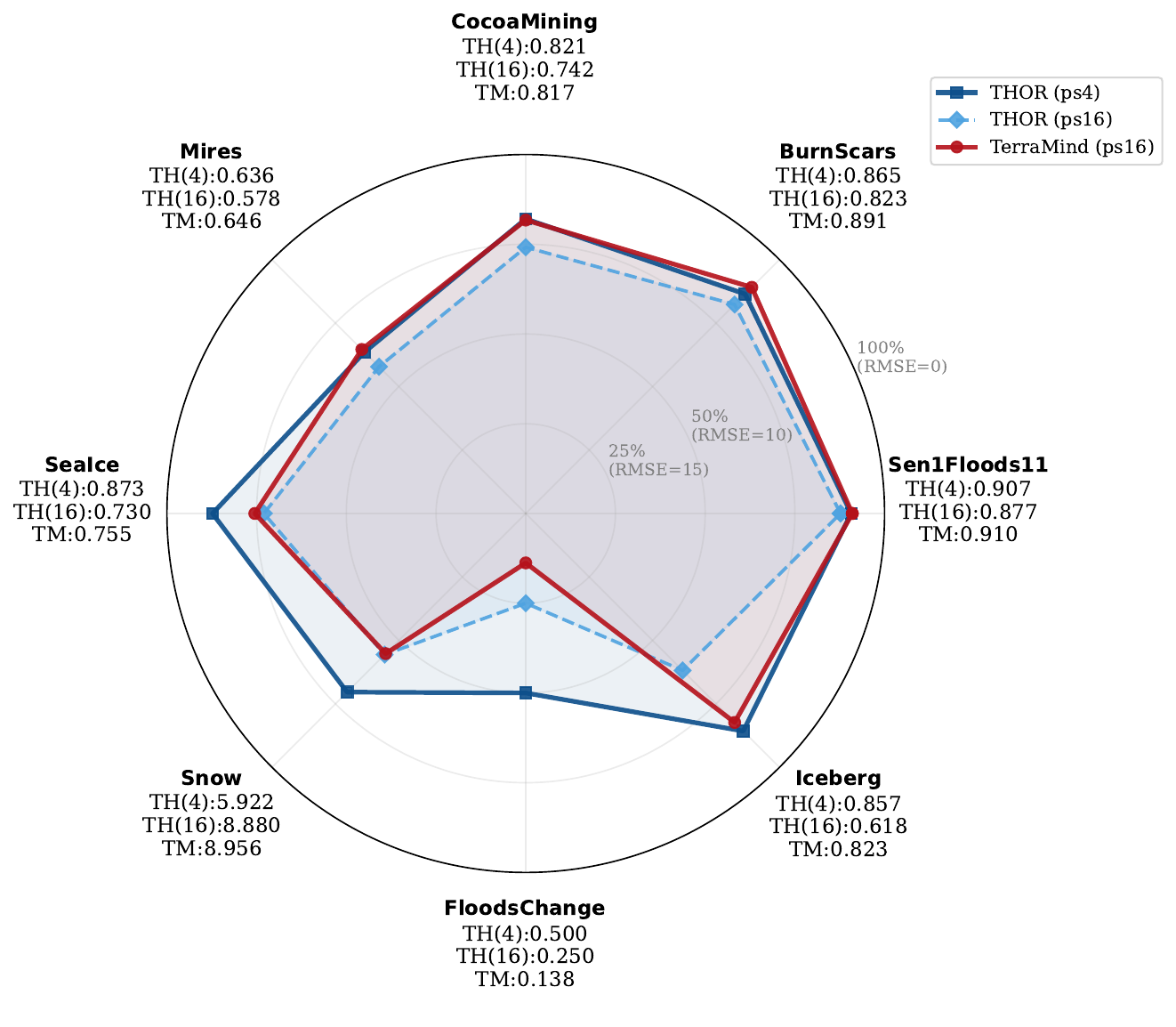}
\caption{%
  \textbf{Radar profile of THOR and TerraMind performance across evaluated tasks.}
  Three profiles are shown at the Base parameter scale: THOR best-across-patch-sizes
  (THOR~(ps4), typically achieved at patch size~4), THOR at patch size~16
  (THOR~(ps16), compute-comparable to TerraMind), and TerraMind
  best-across-configurations (TerraMind~(ps16)). Each axis corresponds to one
  evaluation dataset; raw best values per profile are annotated at each axis
  as TH(4)~/~TH(16)~/~TM.
}
\label{fig:radar}
\end{figure}

\subsection{Overview: Peak Performance and Compute Trade-offs}
\label{sec:overview}

We begin with the headline view. Figure~\ref{fig:pareto} plots test-set
performance against total pipeline cost (encoder + decoder, in GMACs)
across all evaluated use cases and Base-scale configurations. Each
point represents a unique combination of model, decoder, patch size,
and freeze regime.

Two regimes emerge. In the \emph{low-to-moderate compute} envelope
($<$100~GMACs), TerraMind configurations cluster near the Pareto
frontier: its fixed ps16 encoder (${\sim}56$~GMACs) paired with UperNet
achieves 0.910~mIoU on Sen1Floods11, 0.891 on HLS Burn Scars, and
0.646~mIoU on Wetland Mapping -- values that THOR matches or exceeds
only by descending to ps8 or ps4, at $4$--$16\times$ higher encoder
cost. In the \emph{high-compute} regime ($>$200~GMACs), THOR's finer
tokenisation continues to push the frontier, though with diminishing
marginal gains (Section~\ref{sec:patch_size}). This effect is most
pronounced on SAR-based and spatially compact targets: on Flood Zone
Mapping, THOR's best configuration reaches pixel-wise F1\,=\,0.86
against TerraMind's best of 0.33; on Iceberg Detection, THOR leads at
both scales (instance-wise F1\,=\,0.857 vs.\ 0.823 at Base); on Sea Ice
Mapping, THOR reaches 0.873~mIoU against TerraMind's 0.755 (attained
only with TiM); and on Snow Monitoring, THOR's best RMSE (5.92) is
lower than TerraMind's (8.96). Notably, on CocoaMining, THOR
at ps4 (native $128\times128$ input, preserving the 10\,m GSD)
outperforms TerraMind despite TerraMind needing to upsample to
$224\times224$ -- a result we unpack in Section~\ref{sec:anomalies}.
Wetland Mapping is the principal exception to this pattern, and the
only use case where TerraMind's best configuration (0.646~mIoU, frozen
ps16) exceeds THOR's (0.636~mIoU, ps4).

Figure~\ref{fig:radar} complements the Pareto view with a head-to-head
profile of each model's strengths and weaknesses across the evaluated
use cases and summary axes. Across the full set of use cases, no
single model dominates: TerraMind leads on multi-class optical
segmentation and on pure compute efficiency, while THOR's advantage is
concentrated in tasks where spatial granularity or SAR sensitivity
drives discrimination, and where its variable patch size allows it to
trade compute for accuracy in a way that TerraMind, at fixed
resolution, cannot. The remainder of this section unpacks the factors
that give rise to this complementarity.

\subsection{Spatial Tokenisation: The Dominant Performance Lever}
\label{sec:patch_size}

THOR's variable patch size is its primary architectural differentiator.
Halving the patch size quadruples the token count and, approximately,
the self-attention cost. The natural question is whether the resulting
accuracy gains justify this cost, and whether a scaling law governs the
relationship.

Several observations deserve emphasis. First, the
ps32$\to$ps16 transition captures the single largest accuracy
increment on most datasets -- between 2 and 7~pp depending on task and
decoder -- while subsequent halvings contribute progressively less.
Second, the slope is steeper on spatially heterogeneous tasks:
CocoaMining, where artisanal mining sites occupy only a small fraction
of each $128\times128$ patch, benefits most from finer tokenisation (15.5~pp
end-to-end from ps32 to ps4). The FM4CS use cases sharpen this pattern
further. On Flood Zone Mapping and Iceberg Detection -- both
SAR-only tasks with spatially compact, minority-class targets -- coarse
tokenisation is not merely suboptimal but catastrophic: several THOR
decoder configurations collapse to near-zero F1 at ps32 (e.g.\ ps32
context\_aware\_mlp on Flood Zone: F1\,=\,0.004) and recover fully only
at ps4 (F1\,=\,0.86). This corroborates the hypothesis that \emph{tasks
involving rare or spatially compact classes -- such as mining footprints,
flood extents, or iceberg boundaries -- derive disproportionate benefit
from smaller patch sizes}, because coarse tokens risk averaging away the
spectral or backscatter signature of minority-class pixels entirely,
rather than merely degrading it. We revisit this hypothesis in the
CocoaMining analysis, in the Appendix~\ref{sec:anomalies}.

Not every task follows the ps32$\to$ps16-dominant pattern, however. On
Sea Ice Mapping, the largest single-step gain occurs one tier
finer, at ps16$\to$ps8 (+26~pp mIoU, UperNet, FT), with ps32$\to$ps16
contributing comparatively little. Wetland Mapping is
non-monotonic altogether: THOR's ps8 configuration (0.560~mIoU)
underperforms both its coarser (ps16: 0.578) and finer (ps4: 0.636)
neighbours. These exceptions indicate that while finer tokenisation is
directionally beneficial across nearly all tasks, the specific patch
size at which the largest marginal gain occurs is task-dependent rather
than fixed, and the relationship is not always strictly monotonic.


\begin{figure*}[t]
\centering
\includegraphics[width=\textwidth]{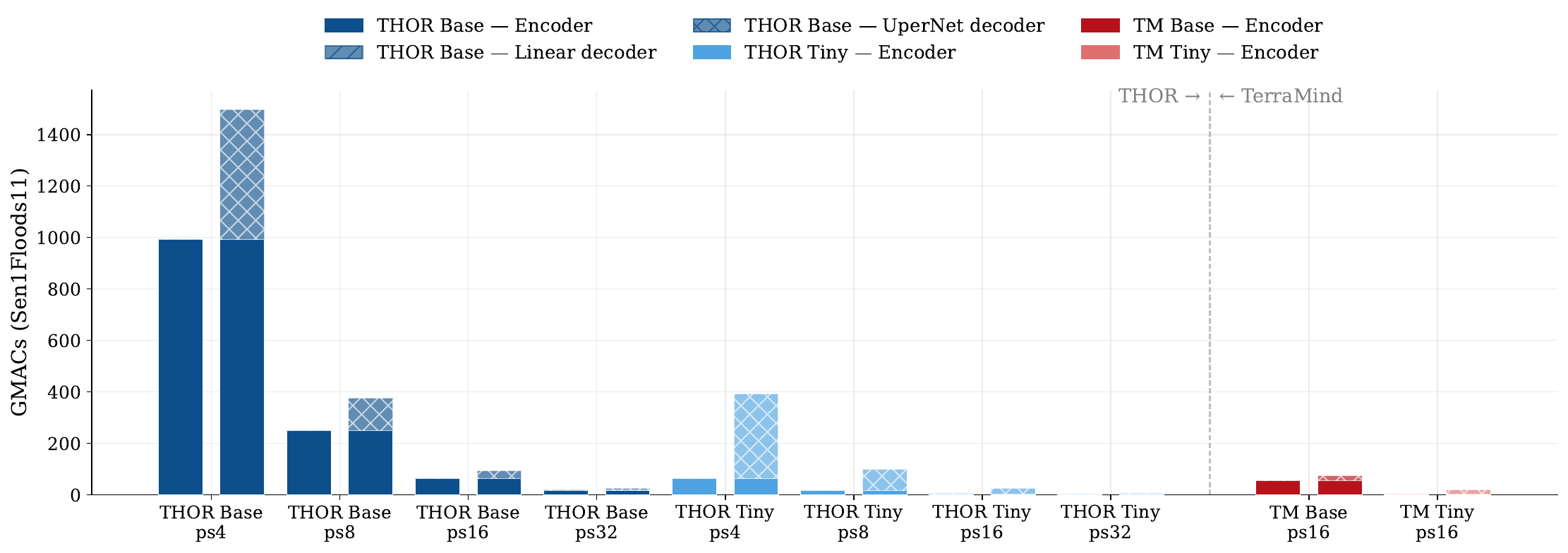}
\caption{%
  \textbf{Encoder and decoder compute cost by model, patch size, and decoder
  type on Sen1Floods11.}
  Grouped bars show total GMACs for each (model, backbone size, patch size)
  configuration, with linear decoder (left bar) and UperNet decoder (right
  bar) shown side by side per group. Bar segments indicate encoder
  compute and decoder compute, stacked
  additively; numbers above bars give total (encoder~+~decoder) GMACs. THOR is shown at patch
  sizes 4, 8, 16, and 32; TerraMind is shown at patch size~16 only, its
  native configuration. 
}
\label{fig:gmacs}
\end{figure*}

\textbf{TerraMind and THOR at matched patch size.}
At matched patch size (ps16, 324~tokens on $288\times288$ crops), THOR-Base
and TerraMind-Base consume nearly identical encoder GMACs (62 vs.\ 56).
Yet TerraMind consistently outperforms THOR at this operating
point -- by 3.3~pp on Sen1Floods11 (UperNet, S2, FT) and 4.5~pp on
HLS Burn Scars, and by a substantially larger margin on Wetland Mapping
(0.646 vs.\ 0.578, 6.8~pp, frozen UperNet). This gap reveals a
fundamental asymmetry: \emph{TerraMind invests computation during
pretraining} (500B tokens, nine modalities, dual-scale loss) \emph{to
produce stronger fixed-resolution features}, while \emph{THOR shifts
the burden to inference time} through finer tokenisation. The reverse
holds on the SAR-only FM4CS use cases: at matched ps16, THOR already
exceeds TerraMind on Sea Ice Mapping before any further patch-size reduction, suggesting that
TerraMind's pretraining advantage is concentrated in optical rather
than SAR representations, probably due to weak consideration of SAR pecularities (e.g. no integration of incidence angle). Both strategies are valid but involve a
trade-off; the optimal choice depends on whether the deployment
constraint is training budget or inference latency, and on the sensor
modality of the target task.

\textbf{Tokens from upsampling vs.\ native resolution.}
An open question is whether THOR's token-count gains can be replicated 
more cheaply by upsampling the input image before patch embedding, 
rather than reducing patch size. We designed two experiments targeting 
this question, reported in full in Appendix~\ref{app:token_resolution}. 
A first experiment on CocoaMining jointly resized train, validation, 
and test images to four target resolutions and observed monotonically 
increasing mIoU. But since the test set is also resized, this confounds 
token count with evaluation-domain alignment and is not a valid 
disentanglement. A second, more controlled experiment on Sen1Floods11 
upsampled only training crops (288--448~px, bilinearly resized to $224\times224$ 
via Albumentations \texttt{Resize}, which uses OpenCV \texttt{INTER\_LINEAR} 
by default) while keeping validation and test at native $512\times512$; via tiled 
inference. Under both frozen and fine-tuned regimes, 
the uptokenized runs degrade slightly relative to the native baseline 
(e.g.\ frozen: 0.905 native vs.\ 0.892--0.899 upsampled; 
fine-tuned: 0.908 vs.\ 0.903--0.905). We therefore conclude that naively 
upsampling training tiles does not replicate the benefit of a genuinely 
finer token grid, and that THOR's ps4/ps8 gains are better attributed 
to \emph{spatially denser feature extraction at native resolution} than 
to an increased token budget per se. Importantly, the CocoaMining 
joint-resize experiment also highlights a benchmark-hygiene risk: resizing 
evaluation tiles can substantially alter reported metrics and must be avoided 
in fair comparisons.

\subsection{Decoder Complexity and Its Interaction with Resolution}
\label{sec:decoder_results}

Section~\ref{sec:patch_size} established that finer tokenisation
improves backbone features. A natural follow-up is whether the
\emph{decoder} can compensate when the backbone operates at coarse
resolution, and conversely, whether an expensive decoder remains
worthwhile when the backbone already provides fine-grained features.
All experiments in this section use UperNet-256
(256~hidden channels, pool scales $\{1,2,3,6\}$, FPN with lateral
connections) and Linear-256 (single projection to 256~channels
followed by learned upsampling), consistent with the PANGAEA evaluation
protocol~\cite{marsocci2024pangaea}, unless otherwise noted.

\begin{figure*}[t]
\centering
\includegraphics[width=\textwidth]{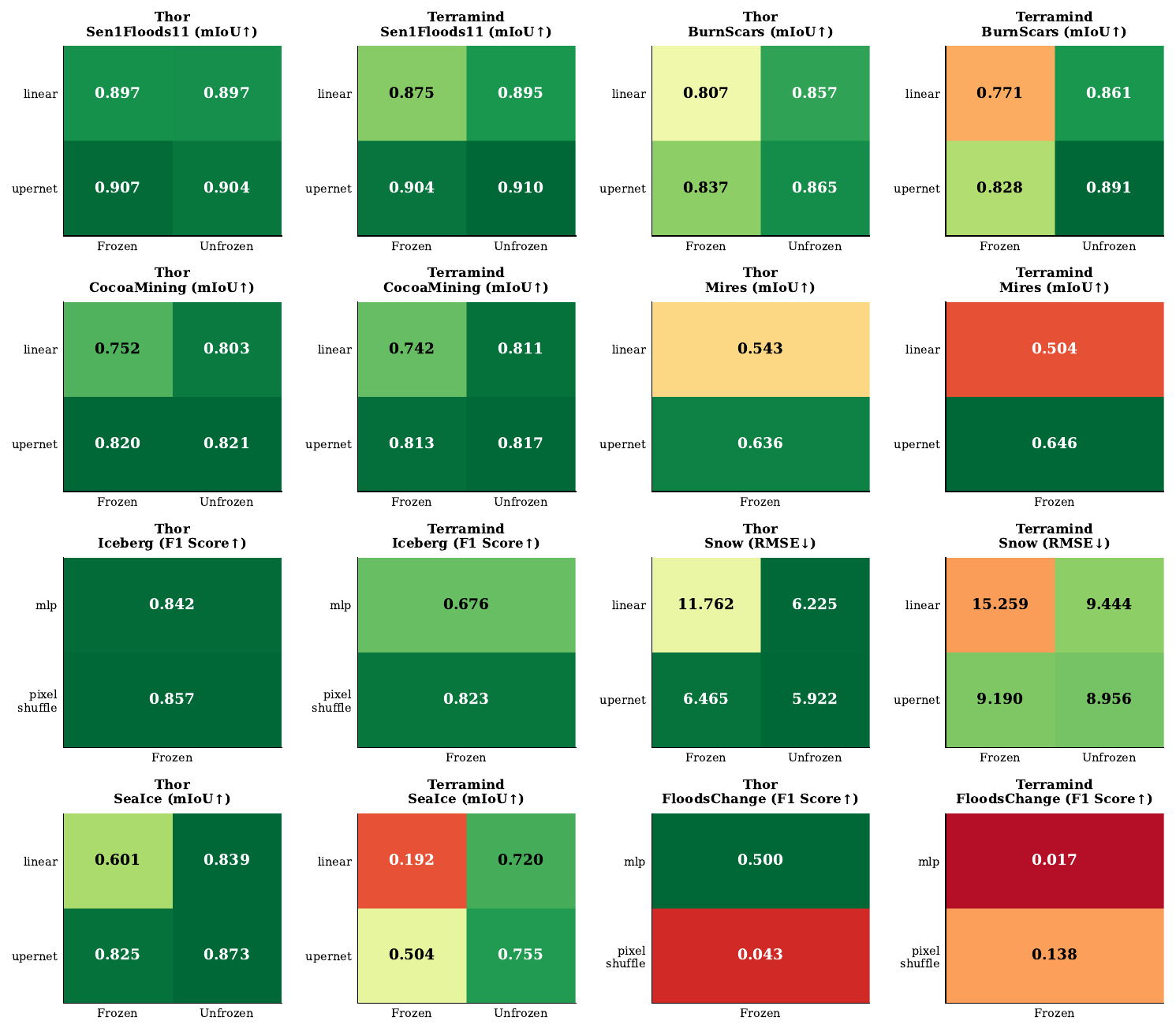}
\caption{%
  \textbf{Decoder and freeze-state interaction on downstream performance,
  by model and dataset.}
  Each heatmap shows performance (best value over configurations) for one
  (model, dataset) pair, with decoder type on the vertical axis and
  encoder freeze-state (Frozen, Unfrozen) on the horizontal axis. Decoders are linear
  and UperNet for mIoU/RMSE datasets, and MLP and PixelShuffle for
  instance-wise F1 datasets. Color scale (diverging, green~=~better,
  red~=~worse) is normalized per dataset to its 2nd--98th percentile
  range across displayed configurations.
}
\label{fig:decoder_freeze}
\end{figure*}

Figure~\ref{fig:decoder_freeze} displays the full interaction between
decoder type, backbone freeze regime, and THOR patch size.

\textbf{THOR: the decoder gap shrinks with resolution.}
On Sen1Floods11 (S1+S2, FT, Base), UperNet leads Linear by
${\sim}$12~pp at ps16 but only ${\sim}$5~pp at ps8. On CocoaMining the
same pattern holds: 3.1~pp at ps16, 1.6~pp at ps8, and 2.4~pp at ps4.
This is consistent with the patchification scaling law reported by Wang
et al.~\cite{wang2025patchscaling}: as patch size decreases, token-level
features already encode sufficient local spatial detail that
task-specific decoder heads become progressively less critical for dense
prediction. In this view, the linear decoder is not inherently
limited -- it benefits from increasingly granular features that reduce
the marginal value of the FPN's multi-scale aggregation. The
decoder-gap convergence is therefore a natural consequence of finer
tokenisation, not an artefact of any particular dataset or training
regime.

Importantly, this pattern was already observed by Forgaard et
al.~\cite{forgaard2026thor} in the original THOR paper. The THOR authors frame
this as: ``the complex decoder was a crutch to compensate for a
`token-starved' encoder.'' Our contribution is to confirm and generalise
this finding across multiple tasks and datasets: within the explored
configuration space and under the present training recipe, the
UperNet--Linear gap narrows consistently at fine patch sizes,
suggesting this is a general property of ViT-based architectures rather
than a dataset-specific effect. On Wetland Mapping, however,
the gap does not fully close: UperNet leads Linear by 9.5~pp at ps16 and
still by 9.3~pp at ps4 (0.636 vs.\ 0.543), the largest residual gap of
any dataset in the ablation. Six-class multi-label segmentation with fine
inter-class boundaries appears to retain a use for multi-scale
aggregation even once the encoder is token-dense, suggesting that
decoder-gap convergence is conditional on task complexity rather than a
universal property of patch size alone.

\textbf{TerraMind: multi-scale features remain essential.}
For TerraMind at fixed ps16, UperNet consistently outperforms Linear.
The gap is large on Sen1Floods11 (8.0~pp, S2 FT) and HLS Burn Scars
(3.0~pp), where tiled inference over $512\times512$ images benefits from the
spatial pyramid's ability to aggregate context across overlapping
windows. On the smaller CocoaMining patches the gap narrows to 0.6~pp,
and on Wetland Mapping it widens again to 14.2~pp (0.646 vs.\ 0.504,
frozen) -- the largest UperNet--Linear gap observed for TerraMind across
any use case, reinforcing that multi-scale aggregation is particularly
valuable for TerraMind on multi-class optical segmentation.

\textbf{The FM4CS decoder heads (MLP, context-aware MLP, PixelShuffle)
tell a less consistent story.} Unlike the UperNet/Linear pair, these
lightweight decoders were evaluated only on Flood Zone Mapping and
Iceberg Detection, and no single decoder dominates across models or
tasks. On Flood Zone Mapping, THOR's MLP and context-aware MLP
decoders far outperform PixelShuffle at every patch size (e.g.\ ps4:
F1\,=\,0.86 and 0.81 vs.\ 0.12), whereas for TerraMind the ranking
inverts -- PixelShuffle is its best decoder (F1\,=\,0.31) while both MLP
variants collapse (F1\,$<$\,0.07). On Iceberg Detection, THOR's
ranking again favours PixelShuffle at fine patch sizes (F1\,=\,0.86 at
ps4, best overall), while MLP decoders remain competitive only at
intermediate ps. This inconsistency indicates that, unlike the
UperNet--Linear pattern, decoder ranking among lightweight heads is not
governed by patch size alone but interacts with model architecture and
task in ways not captured by the token-density argument above; we treat
this as an open question rather than folding it into the general
decoder-gap narrative.

\textbf{Sea Ice Mapping is a partial exception to the general pattern.}
At ps16, THOR's Linear decoder (0.730~mIoU, FT) outperforms its UperNet
counterpart (0.569~mIoU, FT), a reversal not observed on any other
dataset. At ps4 the two decoders converge (0.839 vs.\ 0.873), consistent
with the general narrowing trend, but the sign flip at coarse patch
sizes suggests that for this three-class, floe-boundary task, UperNet's
multi-scale pooling can be actively unhelpful when the backbone is
undertrained on SAR sea-ice imagery, rather than merely providing
diminishing returns.

\textbf{Practical guidance and a key compute-efficiency point.}
A possible interpretation of the shrinking UperNet--Linear gap is that fine patch sizes reduce the need for more expressive decoders. However, Figure~\ref{fig:patchsweep} suggests that this interpretation is incomplete. The total
pipeline cost of ps4~+~Linear on Sen1Floods11 is
$\sim$1{,}004~GMACs, while ps8~+~UperNet costs only
$\sim$375~GMACs -- roughly $2.7\times$ cheaper -- yet achieves competitive
or superior mIoU on most datasets. The UperNet overhead at ps8
($\sim$123~GMACs) is negligible compared with the $4\times$ encoder
saving gained by moving from ps4 to ps8 ($\sim$752~GMACs on $288\times288$
crops). Consequently, ps8~+~UperNet is the recommended operating
point for THOR when compute is constrained. For TerraMind -- or for THOR
at ps16 and coarser, or on tasks such as Wetland Mapping where the
decoder gap does not fully close -- UperNet is consistently worth its
additional cost. When the decoder budget is fixed, \emph{investing in
finer tokenisation yields larger returns than upgrading the decoder
architecture on most tasks}, within the explored configuration space
and under the present training recipe, though this recommendation is
weaker for lightweight FM4CS decoder heads, whose ranking is
task-dependent rather than patch-size-dependent.

\subsection{Frozen vs.\ Fine-Tuned Backbone}
\label{sec:frozen_results}

A central promise of foundation models is that their pretrained
representations should be useful \emph{without} full fine-tuning. We
assess this by comparing frozen-backbone and fine-tuned regimes across
all configurations.

\begin{figure*}[t]
\centering
\includegraphics[width=\textwidth]{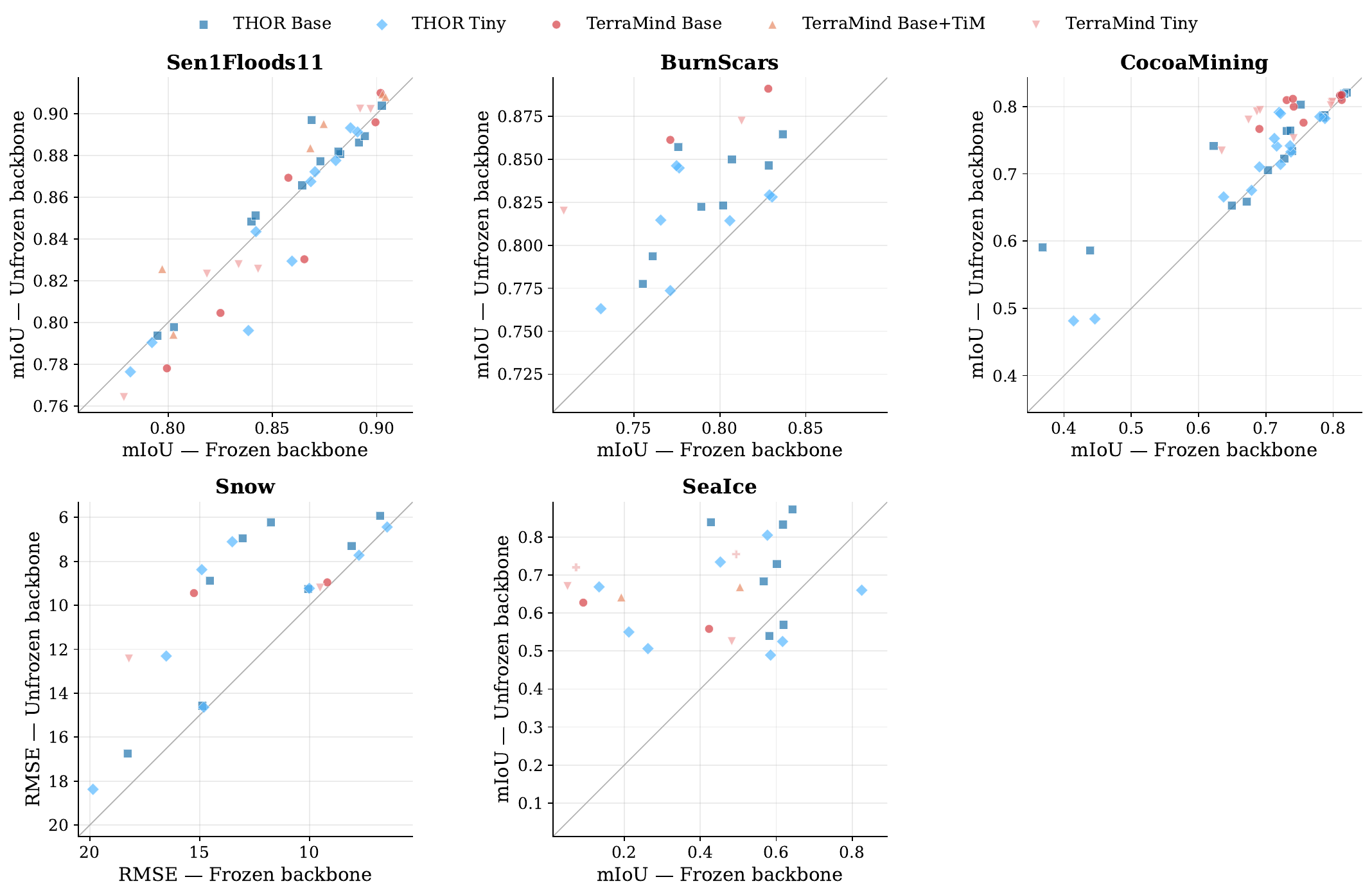}
\caption{%
  \textbf{Frozen versus fine-tuned backbone performance across datasets,
  decoders, and patch sizes.}
  Each panel plots performance with a frozen backbone (x-axis) against
  performance with a fine-tuned (unfrozen) backbone (y-axis) for one
  dataset; each point is one (model, version, modality, decoder, patch
  size) configuration for which both freeze states are available.
  Diagonal line indicates equal performance; points above the diagonal
  indicate that fine-tuning improves performance, points below indicate
  that the frozen backbone performs better.
}
\label{fig:frozen}
\end{figure*}

\textbf{TerraMind: fine-tuning provides a consistent but modest lift.}
Across all datasets and decoders, fine-tuning improves over freezing by
1--8~pp. The largest gains appear with linear decoders. On Sen1Floods11
(S2, UperNet), fine-tuning adds only 0.8~pp (0.910 vs.\ 0.902),
suggesting that TerraMind's pre-trained features are already strongly
aligned with flood-mapping semantics (probably also because it was pretrained on LULC). On HLS Burn Scars the gap widens to
6.3~pp, reflecting a larger domain shift from pretraining.

\textbf{THOR: fine-tuning generally helps, with one notable exception.}
On most configurations, fine-tuning provides 2--5~pp improvement.
However, on Sen1Floods11 at ps8, the \emph{frozen} backbone (0.894,
UperNet, S2) marginally outperforms the fine-tuned counterpart (0.889).
A practical hardware constraint is relevant here: at ps4, fitting a
batch of~8 samples in 46~GB of VRAM is not feasible during full
fine-tuning, so we used a physical batch size of~2 with $4\times$
gradient accumulation (effective batch~8). To verify that this does not
drive the anomaly, we ran matched experiments at physical batch size~2
for both frozen and fine-tuned regimes; the frozen--FT gap direction and
magnitude are reproduced in both conditions, suggesting the anomaly is
not purely a batch-size artefact. 

\textbf{Sea Ice Mapping reveals a patch-size-dependent reversal for
THOR.} At fine patch sizes, fine-tuning provides a large and consistent
gain: at ps4, FT reaches 0.873~mIoU against 0.642 frozen (+23~pp,
UperNet); at ps8, 0.833 against 0.618 frozen (+22~pp). At coarse patch
sizes the direction inverts, mirroring the Sen1Floods11 ps8 anomaly but
more strongly: at ps16, the frozen backbone (0.619) exceeds fine-tuned
(0.569, $-$5~pp), and at ps32 frozen again leads (0.581 vs.\ 0.540,
$-$4~pp). Taken together with the Sen1Floods11 exception, this suggests
that fine-tuning a token-starved (coarse-patch) backbone can actively
degrade features relative to leaving them frozen, while at fine patch
sizes -- where the encoder already has ample spatial resolution to
adapt -- fine-tuning is unambiguously beneficial.

\textbf{TerraMind's frozen features do not always track its fine-tuned
ceiling.} On Sea Ice Mapping, TerraMind's frozen backbone collapses
sharply relative to its fine-tuned counterpart, with the linear decoder
falling to near-random performance (0.092~mIoU frozen vs.\ 0.627
fine-tuned, ps16) and UperNet showing a similarly large gap (0.423
frozen vs.\ 0.558 fine-tuned). This is the largest frozen--FT gap
observed for TerraMind across any use case, and stands in contrast to
the modest, consistent lift reported above -- indicating that TerraMind's pre-trained representations,
despite including an S1 modality, do not transfer as readily to
SAR-only sea-ice segmentation as they do to optical or S1+S2 fusion
tasks.

\textbf{Coverage note.} Owing to computational constraints and use case specific requirements, not every ablation could be performed for every dataset. Flood Zone Mapping, Iceberg Detection, and Wetland Mapping were evaluated only under the frozen backbone regime in the present ablation. Consequently, we are unable to report a frozen versus fine-tuned comparison for these use cases.

\subsection{Multimodal Input Composition}
\label{sec:multimodal_results}

A commonly held assumption in the EO community is that fusing
complementary sensors -- optical and SAR, in particular -- should improve
performance over either modality alone. Our ablation on Sen1Floods11 and
CocoaMining examines this assumption in the full-data regime.

\begin{figure*}[t]
\centering
\includegraphics[width=\textwidth]{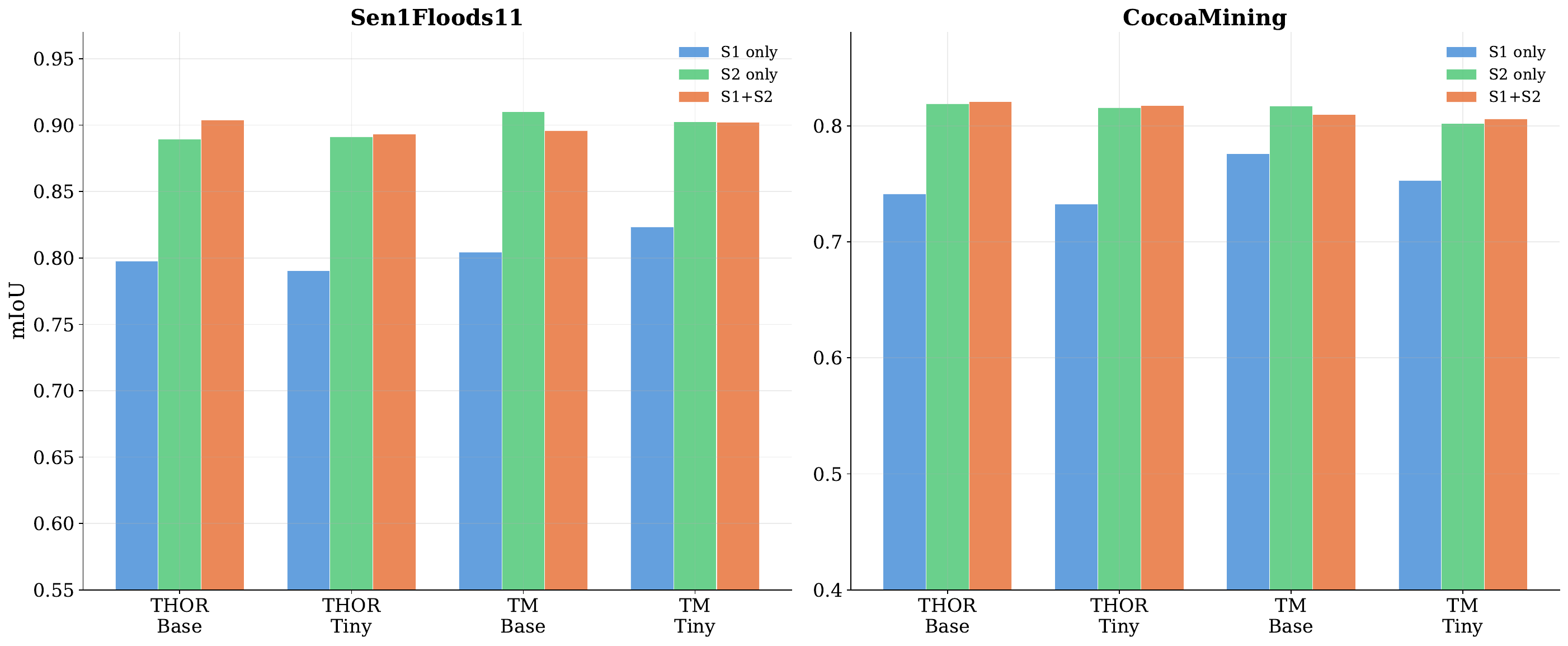}
\caption{%
  \textbf{Modality ablation on Sen1Floods11 and CocoaMining.}
  Grouped bars show best mIoU per model, backbone size, and input
  modality (S1~only, S2~only, S1+S2 fusion), for the UperNet decoder
  with unfrozen backbone.
}
\label{fig:modality}
\end{figure*}

\textbf{S2 is the dominant modality.} On Sen1Floods11, TerraMind's best
S2-only result (0.910, UperNet FT) \emph{exceeds} its best S1+S2 result
(0.896) by 1.4~pp. THOR follows the same pattern at ps8: S2 UperNet
frozen (0.894) vs.\ S1+S2 (0.891). On CocoaMining, the margin is
smaller but the direction holds.

\textbf{Simple S1+S2 fusion schemes did not reliably outperform S2-only
on the tested datasets, in the full-data regime.}
This negative result implicates the \emph{fusion mechanism} rather than
the data: both THOR's early concatenation and TerraMind's per-modality
projection with mean pooling are relatively shallow fusion strategies.
More expressive cross-attention or gating mechanisms may be needed to
unlock SAR complementarity. Importantly, this result is data-regime
dependent: both the original THOR paper~\cite{forgaard2026thor}
(Table~3: S1+S2\,=\,87.70 vs.\ S2\,=\,87.25 on Sen1Floods11 at 10\%
training data) and the original TerraMind paper~\cite{jakubik2025terramind}
(Table~7: S1+S2\,=\,90.62 vs.\ S2\,=\,89.57) report a modest but
positive SAR contribution in the low-data or multi-dataset regime. The
discrepancy is consistent with a data-regime effect: when optical signal
is abundant (100\% training data), the marginal information from SAR may
be insufficient to overcome the noise introduced by shallow fusion.

\textbf{S1-only is informative for flood mapping, insufficient for
spectral tasks.} On Sen1Floods11, S1-only performance is non-trivial
(THOR 0.803, TerraMind 0.825), reflecting the typical SAR backscatter
signature of water bodies (i.e low backscatter). On CocoaMining, where distinguishing cocoa
from the background class requires spectral information, S1-only drops
sharply.

\textbf{DEM adds no measurable benefit.} On CocoaMining, TerraMind's
S1+S2+DEM configuration matches S2-only, consistent with the relatively
flat topography of southern Ghana.

\subsection{Snow Monitoring: Cross-Sensor Generalisation}
\label{sec:cross_sensor}

The Snow use case uses Sentinel-3 SLSTR imagery (${\sim}500$\,m GSD), a
sensor absent from TerraMind's pretraining corpus, unlike THOR which
natively unifies S1/S2/S3. We evaluate TerraMind by mapping SLSTR bands
onto the closest Sentinel-2 wavelengths and feeding them through its
existing S2 input head -- a genuine cross-sensor \emph{and}
cross-resolution shift, since TerraMind saw S2 only at 10--20\,m during
pretraining. Despite this, TerraMind remains competitive
(Table~\ref{tab:full_results_merged}), suggesting
that spectral compatibility with a pretrained input head can matter more
than native spatial resolution, and that wavelength-based band mapping
is a viable route to extending a GFM beyond its native sensor suite
without retraining a dedicated input head.

\subsection{Model Scale: Tiny vs.\ Base}
\label{sec:scale_results}

\begin{figure*}[t]
\centering
\includegraphics[width=\textwidth]{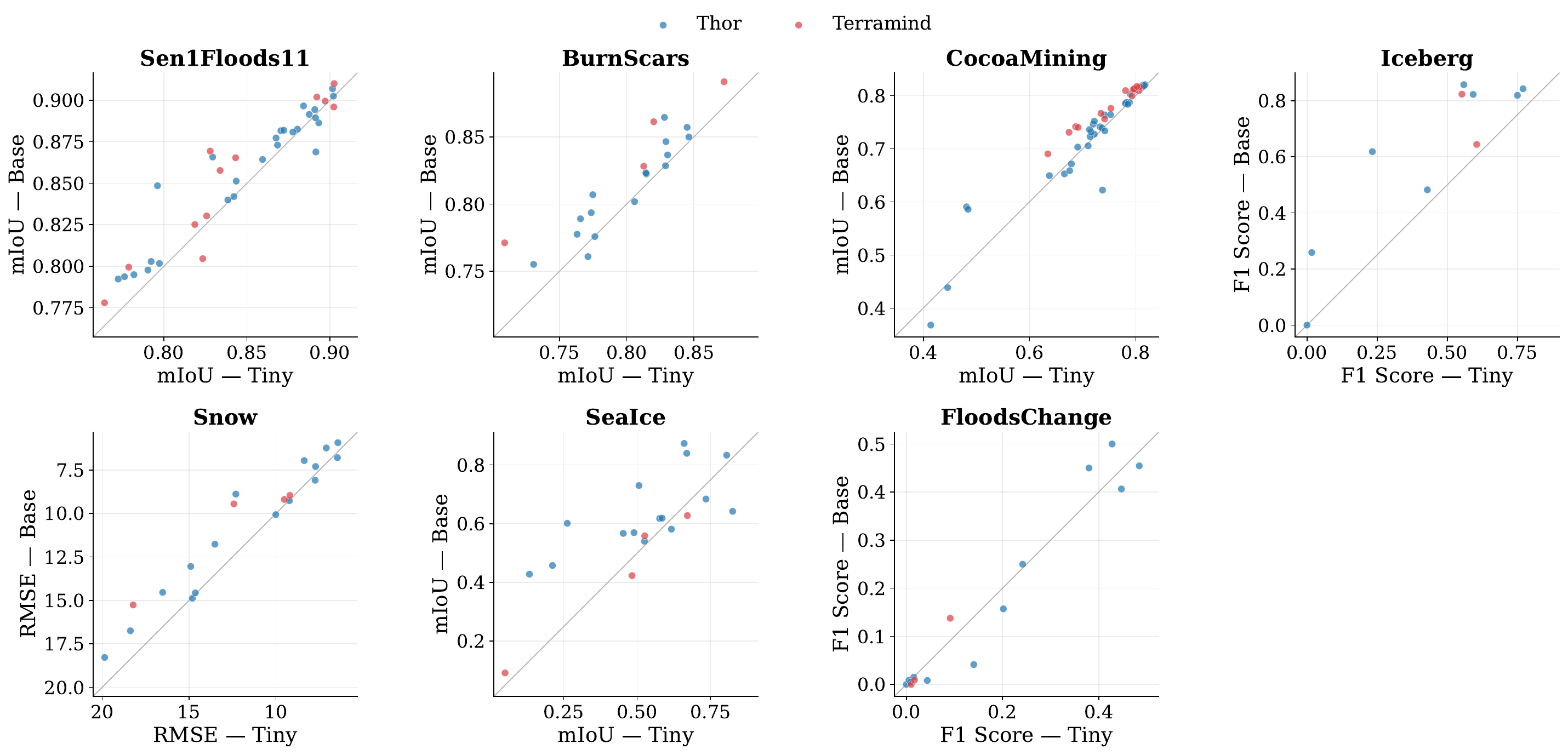}
\caption{%
  \textbf{Base versus Tiny backbone scaling across datasets, decoders,
  and patch sizes.}
  Each panel plots per-configuration performance at Tiny scale (x-axis)
  against Base scale (y-axis) for one dataset, matched across model,
  modality, decoder, freeze-state, and patch size. Diagonal line
  indicates equal performance; points above the diagonal indicate that
  Base outperforms Tiny.
}
\label{fig:scale}
\end{figure*}

Both THOR and TerraMind are evaluated at Tiny and Base scales.
Figure~\ref{fig:scale} compares the two scales across datasets,
decoders, and -- for THOR -- patch sizes.

\textbf{Base consistently outperforms Tiny, but the gap is task-dependent.} On Sen1Floods11 (S2, UperNet,
FT), TerraMind-Base leads Tiny by 0.8~pp. On HLS Burn Scars the gap
widens to 1.9~pp; on CocoaMining to 1.5~pp. For THOR at ps8, the gap is
smaller -- only 0.2~pp on Sen1Floods11 -- suggesting that fine
tokenisation partially compensates for reduced model capacity.

\textbf{The FM4CS use cases largely confirm this pattern, with the
scale gap sometimes amplified.} On Iceberg Detection, Base leads Tiny
for both models, and more sharply than on the datasets: THOR by
8.7~pp (instance-wise F1 0.857 vs.\ 0.770) and TerraMind by 21.9~pp
(0.823 vs.\ 0.604). On Flood Zone Mapping the gap is small and
consistent with the magnitude (THOR context-aware MLP,
ps4, frozen: 0.709 Base vs.\ 0.680 Tiny). Wetland Mapping was evaluated
at Base scale only, so no scale comparison is available for this use
case.

\textbf{Sea Ice Mapping is the principal exception to Base~$>$~Tiny.}
At ps4 with a frozen backbone, THOR-Tiny (0.825~mIoU) substantially
outperforms THOR-Base (0.642~mIoU) -- an 18.3~pp reversal, the largest
scale inversion observed in the benchmark. The ordering reverts to the
expected direction once the backbone is fine-tuned (Base 0.873 vs.\
Tiny 0.660), indicating the reversal is specific to the frozen regime
rather than a property of Tiny features generally. TerraMind shows a
smaller but analogous reversal at ps16, frozen: Tiny (0.483~mIoU)
exceeds Base (0.423~mIoU), and the reversal persists with TiM applied
(Tiny\_TiM 0.755 vs.\ Base\_TiM 0.668, a 8.7~pp gap in Tiny's favour).
A plausible explanation is that Base-scale pre-trained features do not
always transfer better than Tiny-scale features when the target sensor
modality is under-represented in pretraining -- i.e.\ that capacity
alone does not guarantee more transferable frozen representations
outside the sensor domains a model was principally trained on.
However, as discussed in Appendix~\ref{sec:anomalies}, Sea Ice
Mapping's train/val/test splits show signs of distributional
instability, which offers an equally plausible, split-driven account
of these reversals; given the limited number of runs per configuration,
we cannot adjudicate between the two, and refrain from strongly
inferring the capacity-transfer explanation from this dataset alone.

\textbf{Tiny models are disproportionately sensitive to decoder
choice.} For TerraMind-Tiny on HLS Burn Scars, the UperNet--Linear gap
is 5.2~pp (vs.\ 3.0~pp for Base). The weaker backbone produces less
spatially structured features, increasing the decoder's marginal
contribution. This effect is consistent with the wider UperNet--Linear
gaps observed for TerraMind on Wetland Mapping (Section~\ref{sec:decoder_results}),
though that comparison is confined to Base scale in the present ablation.

\textbf{Cost--accuracy trade-off.} TerraMind-Tiny's encoder costs
${\sim}15\times$ less than Base (3.7 vs.\ 56.0~GMACs), yet reaches
94--99\% of Base's mIoU with UperNet, and
in some Sea Ice Mapping configurations matches or exceeds Base outright
at a fraction of the cost. Combined with the Pareto analysis of
Figure~\ref{fig:pareto}, \emph{TerraMind-Tiny + UperNet} emerges as the
most compute-efficient configuration in our benchmark. However, the
decoder cost (${\sim}$16.7~GMACs for UperNet) does not scale with
encoder size, meaning the decoder dominates the total pipeline at Tiny
scale.
\subsection{Low-Data Regime}
\label{sec:low_data}

\begin{figure*}[t]
\centering
\includegraphics[width=\textwidth]{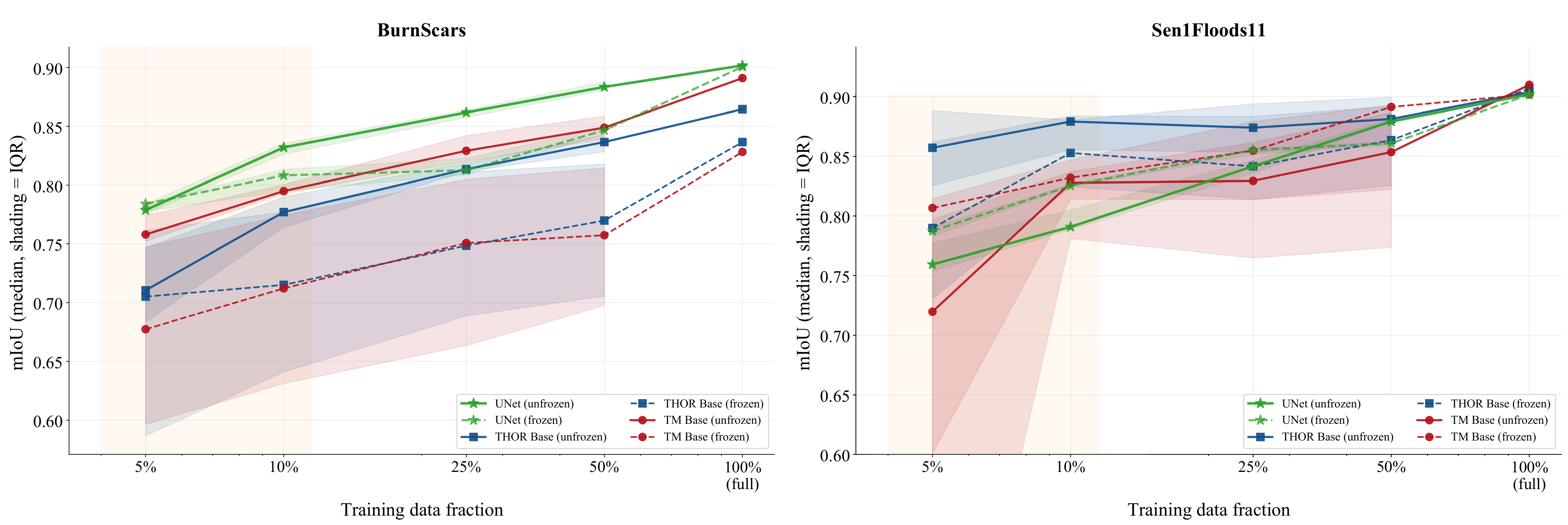}
\caption{%
  \textbf{Data-efficiency curves on Sen1Floods11 (left) and HLS Burn
  Scars (right).}
  Mean mIoU over 3--6 seeds as a function of training data fraction
  (5, 10, 25, 50, 100\%; log scale); shaded bands show $\pm1$ standard
  deviation. UNet uses a ResNet-50 backbone; THOR and TerraMind use the
  UperNet decoder.
}
\label{fig:data_efficiency}
\end{figure*}

Figure~\ref{fig:data_efficiency} and Tables~\ref{tab:low_data_frozen}--\ref{tab:low_data_ft}
report mean mIoU across four training fractions (3 seeds per
configuration, UperNet decoder) for THOR and TerraMind alongside a
task-specific UNet baseline (see Appendix~\ref{app:data_efficiency} for
extended analysis).

\begin{table}[t]
\centering
\caption{%
  Low-data mIoU  --  \textbf{frozen backbone}, UperNet decoder.
  Image counts: S1F 13/26/63/126; BRS 25/51/130/262.
  \textbf{Bold}~=~best per (dataset, fraction).}
\label{tab:low_data_frozen}
\resizebox{\columnwidth}{!}{%
\begin{tabular}{lrrrrrrrr}
\toprule
 & \multicolumn{4}{c}{\textbf{Sen1Floods11}} & \multicolumn{4}{c}{\textbf{HLS Burn Scars}} \\
\cmidrule(lr){2-5}\cmidrule(lr){6-9}
\textbf{Config} & 5\% & 10\% & 25\% & 50\% & 5\% & 10\% & 25\% & 50\% \\
\midrule
TM ps16       & 0.820 & 0.846 & 0.877 & \textbf{0.893} & 0.752 & 0.779 & 0.807 & 0.815 \\
THOR ps4      & \textbf{0.864} & \textbf{0.884} & \textbf{0.882} & 0.892 & \textbf{0.764} & 0.785 & 0.811 & 0.818 \\
THOR ps16     & 0.829 & 0.839 & 0.852 & 0.861 & 0.703 & 0.737 & 0.777 & 0.796 \\
\midrule
UNet (ref.)   & 0.777 & 0.812 & 0.847 & 0.871 & \textbf{0.780} & \textbf{0.817} & \textbf{0.838} & \textbf{0.865} \\
\bottomrule
\end{tabular}%
}
\end{table}

\begin{table}[t]
\centering
\caption{%
  Low-data mIoU  --  \textbf{fine-tuned (FT) backbone}, UperNet decoder.
  Image counts: S1F 13/26/63/126; BRS 25/51/130/262.
  \textbf{Bold}~=~best per (dataset, fraction).}
\label{tab:low_data_ft}
\resizebox{\columnwidth}{!}{%
\begin{tabular}{lrrrrrrrr}
\toprule
 & \multicolumn{4}{c}{\textbf{Sen1Floods11}} & \multicolumn{4}{c}{\textbf{HLS Burn Scars}} \\
\cmidrule(lr){2-5}\cmidrule(lr){6-9}
\textbf{Config} & 5\% & 10\% & 25\% & 50\% & 5\% & 10\% & 25\% & 50\% \\
\midrule
TM ps16       & 0.813 & 0.842 & 0.861 & 0.890 & 0.756 & 0.802 & \textbf{0.842} & 0.862 \\
THOR ps4      & \textbf{0.888} & \textbf{0.882} & \textbf{0.896} & \textbf{0.899} & 0.746 & 0.794 & 0.810 & 0.827 \\
THOR ps16     & 0.814 & 0.848 & 0.857 & 0.853 & 0.700 & 0.749 & 0.786 & 0.813 \\
\midrule
UNet (ref.)   & 0.777 & 0.812 & 0.847 & 0.871 & \textbf{0.780} & \textbf{0.817} & 0.838 & \textbf{0.865} \\
\bottomrule
\end{tabular}%
}
\end{table}

\textbf{Frozen regime (Table~\ref{tab:low_data_frozen}).}
With frozen backbone, THOR ps4 leads on Sen1Floods11 at 5--25\%,
while TerraMind ps16 edges ahead at 50\% (0.893 vs.\ 0.892). On
HLS Burn Scars, UNet dominates at every fraction; both GFMs trail
by up to 11.4\,pp. Two complementary factors likely contribute:
(i)~the task is purely optical with no SAR component, eliminating
any multimodal pretraining advantage; (ii)~the HLS (Harmonized
Landsat Sentinel-2) product cross-calibrates Landsat and Sentinel-2
acquisitions, introducing a source distribution shift relative to the
raw Sentinel-2 pretraining data of both GFMs -- a setting analogous to
the DeepGlobe$\to$DFC2022 source shift in EarthShift~\cite{doerksen2026earthshift},
where GFMs showed no robustness advantage over task-specific supervised
models. A compact UNet is unaffected by pretraining distribution
mismatch and benefits directly from the task-aligned training signal.

\textbf{Fine-tuned regime (Table~\ref{tab:low_data_ft}).}
Unlocking the backbone amplifies THOR ps4 considerably on
Sen1Floods11: at 5\% (13 images) it achieves
mIoU\,=\,0.888. This is equivalent to what UNet reaches at 50\% (126 images, 0.871)
at a ${\sim}10\times$ data reduction. THOR ps4 leads
at 5, 10 and 25\% fractions. On Burn Scars the ordering is unchanged: UNet
wins at 5\%, 10\%, and 50\%; TerraMind FT edges ahead at 25\%
(0.842 vs.\ 0.838). Overall, the relative efficiency advantage is dataset-dependent and reflects an interaction between pretraining coverage, sensor modality, and spatial resolution. In our experiments, TerraMind is particularly competitive on optical and multimodal use cases, while THOR benefits more when finer spatial tokenisation is needed.
\subsection{Robustness and Sensitivity Analysis}
\label{sec:robustness}

The preceding subsections focus on Sen1Floods11, HLS Burn Scars,
CocoaMining, and Snow Monitoring. We exclude the remaining FM4CS use
cases from this closing analysis for dataset-specific reasons: Wetland
Mapping was evaluated with a substantially reduced ablation grid owing
to its large per-run training time, and Flood Zone Mapping and Sea Ice
Mapping exhibit the split instability and threshold-like behaviour
discussed in Appendix~\ref{sec:anomalies}, which limits their
suitability as a basis for a clean, cross-model sensitivity
decomposition. We now ask: \emph{across the full ablation grid on
these four datasets, which factors explain the most performance
variance?}

\begin{figure*}[t]
\centering
\includegraphics[width=\textwidth]{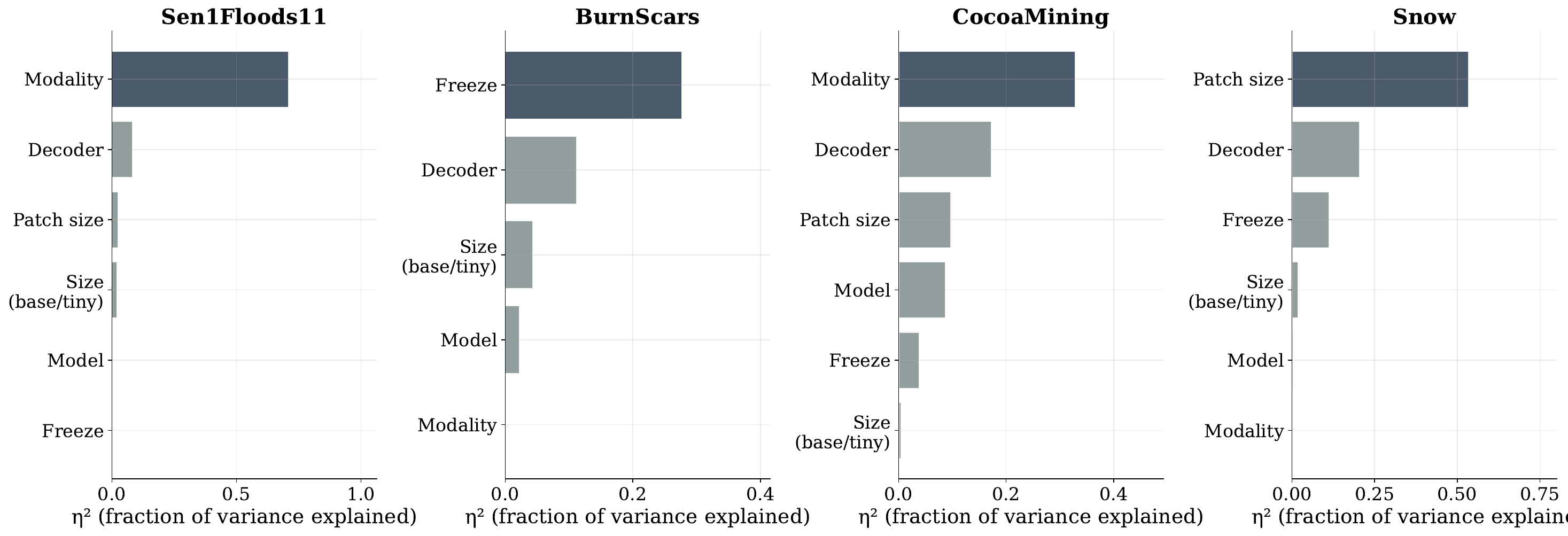}
\caption{%
  \textbf{Variance decomposition of downstream performance by design
  factor, across four datasets.}
  Horizontal bars show $\eta^2$ (fraction of variance explained) from
  one-way ANOVA of each design factor (model, backbone size, decoder,
  freeze-state, input modality, and patch size, where applicable) against
  the dataset's performance metric, ranked by descending $\eta^2$ within
  each panel.
}
\label{fig:variance}
\end{figure*}

To quantify which design factor explains the largest share of the observed performance variation, we use a one way ANOVA based decomposition. ANOVA, short for analysis of variance, partitions the total variability in mIoU into variation explained by differences between the levels of a factor and residual variation within each level. We report this effect size using
\[
\eta^2 = \frac{SS_{\mathrm{between}}}{SS_{\mathrm{total}}},
\]
which measures the fraction of total variance attributable to that factor. Figure~\ref{fig:variance} presents this variance decomposition across all
THOR and TerraMind runs on the FAST-EO datasets, showing which design factor accounts for the largest share of the mIoU variation across our controlled ablations.

The decomposition yields a key finding: \textit{within the explored
configuration space, patch size and decoder
type jointly explain more performance variance than model identity}.
An EO practitioner choosing between THOR and TerraMind should first
determine their compute budget and decoder constraints, as these choices
will have a larger impact on downstream performance than the foundation
model itself -- at least under the present training recipe and on the
evaluated task collection.



Additional robustness diagnostics -- including rank consistency across
datasets (Figure~\ref{fig:rank_consistency}), compute-normalised
rankings (Figure~\ref{fig:compute_ranking}), and marginal value
decomposition (Figure~\ref{fig:marginal_value}) -- are provided in
Appendix~\ref{app:robustness}.

\subsection{Hyperspectral Use Cases}
\label{sec:hyperspectral}

Both FAST-EO hyperspectral use cases -- Methane Leaks and
HYPERVIEW -- involve airborne sensors with hundreds of narrow spectral
bands, a modality that differs fundamentally from the broadband
satellite sensors used during pretraining of both THOR and TerraMind.
A relevant asymmetry applies for Sen1Floods11 (L1C, 13 bands): TerraMind ingests all 13 bands natively while THOR drops the three L1C-only bands (see Section~\ref{sec:usecases}). For the airborne hyperspectral inputs here, both models are reduced to a Sentinel-2-matched subset regardless, so the L1C/L2A distinction does not further affect these two use cases.
Table~\ref{tab:hyperspectral_results} reports the key results.

\begin{table}[t]
\centering
\caption{Hyperspectral use case results. UC2 Methane Leaks: Overall Accuracy~($\uparrow$); UC4 HYPERVIEW: RMSE~($\downarrow$, lower is better). FT~=~full fine-tuning; Fr~=~frozen backbone. \textbf{Bold}~=~best per use case.}
\label{tab:hyperspectral_results}
\resizebox{\columnwidth}{!}{%
\begin{tabular}{lllll}
\toprule
\textbf{Model} & \textbf{Backbone} & \textbf{Decoder} & \textbf{ps} & \textbf{Metric} \\
\midrule
\multicolumn{5}{l}{\textit{UC2  --  Methane Leaks (OA $\uparrow$)}} \\
\textbf{THOR Base}  & Fr & UperNet & 8  & \textbf{0.761} \\
TerraMind Base      & Fr & UperNet & 16 & 0.728 \\
\midrule
\multicolumn{5}{l}{\textit{UC4  --  HYPERVIEW (RMSE $\downarrow$)}} \\
THOR Base           & FT & UperNet & 4  & 1.037 \\
\textbf{TerraMind Base} & FT & UperNet & 4 & \textbf{0.965} \\
\bottomrule
\end{tabular}%
}
\end{table}

\subsubsection{Methane Leaks}
\label{sec:methane_results}

THOR at ps8 with frozen backbone and UperNet decoder achieves the best result (OA~=~0.761), outperforming the best frozen TerraMind configuration (OA~=~0.728). Two observations stand out. First, \textit{frozen backbones outperform fine-tuned ones}
on both models. This may appear counterintuitive -- fine-tuning should help the backbone
adapt to the new modality -- but resolves when considering dataset size and shift
magnitude. With a small annotated set and extreme spectral domain shift (432 airborne
bands vs.~broadband satellite pretraining), fine-tuning risks overwriting
satellite-derived spatial priors (edge detection, texture, structural patterns) that
remain partially useful for plume segmentation, while simultaneously overfitting to
narrow hyperspectral statistics. Doerksen and
Kerner~\cite{doerksen2026earthshift} report a consistent pattern across GFMs: encoder
fine-tuning does not improve, and sometimes worsens, distributional robustness under
sensor and source shifts. The contrast with the HYPERVIEW use case -- where fine-tuning \emph{does}
outperform frozen (0.965 vs.~competitive frozen baselines) -- supports a
dataset-size explanation: HYPERVIEW provides 1{,}732 training patches, enough for the backbone to adapt. Second, \textit{ps8 outperforms other patch sizes} for THOR,
suggesting methane plume structures are resolved at intermediate spatial scales without
incurring the overhead of ps4.

\subsubsection{HYPERVIEW (Soil Property Regression)}
\label{sec:hyperview_results}

TerraMind at ps4 with full fine-tuning achieves the best RMSE~(0.965), marginally outperforming THOR at the same patch size and regime~(1.037). As with Methane Leaks, frozen backbones remain competitive, which we attribute to the large spectral domain gap between the HYPERVIEW airborne sensor and Sentinel-based pretraining rather than to a small dataset (HYPERVIEW provides 1{,}732 training patches). The ps4 advantage for THOR is consistent with the general token-scaling pattern of Section~\ref{sec:patch_size}: finer tokenisation compensates for reduced spectral coverage, at the cost of substantially higher compute.

\subsection{TerraMind at Patch Size 4}
\label{sec:terramind_ps4}

The comparisons in the preceding sections rest on one important
asymmetry: THOR at ps4 and TerraMind at its native ps16 are not
directly comparable in either performance or compute, since one
architecture's pretraining explicitly targets variable, fine-grained
tokenisation while the other's does not. Conversely, TerraMind at
ps16 is THOR's \emph{native} operating point only in the sense that
ps16 is one option among several THOR was pretrained to support;
THOR's own architectural philosophy anticipates operation at coarser
patch sizes as well, so neither model is being evaluated purely
outside its intended design space, but the two are still not tested
on equal architectural footing at any single patch size.

To probe this asymmetry from an additional angle, we obtained, in
cooperation with IBM Research, a version of TerraMind tiny, whose pixel-level
patch embedding was \emph{partially} re-pretrained at patch size~4. The
embedding layer was retrained for 48 epochs on the TerraMesh dataset;
the tokenizer was left untouched and no validation split was used
during this retraining, both due to computational constraints. This is
therefore not a full re-pretraining under TerraMind's original recipe
(500B tokens, nine modalities, dual-scale loss), but a targeted probe
of whether TerraMind's architecture can be adapted, even partially, to
finer tokenisation. As this variant is not part of any official
release, we treat it as supplementary to the main ablation
grid rather than central to it, and restrict evaluation to the FAST-EO datasets (Sen1Floods11, HLS Burn Scars, CocoaMining) at
Tiny scale. Full pretraining details and additional results are
provided in Appendix~\ref{app:tmps4results_results}.

\begin{table}[t]
\centering
\caption{%
  TerraMind-ps4 (partially re-pretrained, Tiny scale) mIoU across
  decoder and freeze regime. FT~=~fine-tuned; Fr~=~frozen backbone.
  Modality: S1+S2 for Sen1Floods11, S2 for HLS Burn Scars,
  S1+S2+DEM for CocoaMining.}
\label{tab:terramind_ps4}
\begin{tabular}{lcccc}
\toprule
\textbf{Dataset} & \textbf{Decoder} & \textbf{Regime} & \textbf{mIoU} \\
\midrule
\multirow{4}{*}{Sen1Floods11}
  & UperNet & FT & \textbf{0.919} \\
  & UperNet & Fr & 0.915 \\
  & Linear  & FT & 0.881 \\
  & Linear  & Fr & 0.825 \\
\midrule
\multirow{4}{*}{HLS Burn Scars}
  & Linear  & FT & \textbf{0.894} \\
  & UperNet & FT & 0.892 \\
  & UperNet & Fr & 0.845 \\
  & Linear  & Fr & 0.680 \\
\midrule
\multirow{4}{*}{CocoaMining}
  & UperNet & FT & \textbf{0.830} \\
  & UperNet & Fr & 0.805 \\
  & Linear  & FT & 0.808 \\
  & Linear  & Fr & 0.639 \\
\bottomrule
\end{tabular}
\end{table}

\begin{table}[t]
\centering
\caption{%
  Best mIoU per dataset: TerraMind-ps4 (retrofit, Tiny) vs.\ the three
  reference points discussed throughout this paper. TerraMind-ps16 and
  THOR-ps4 are Base-scale; the retrofit is Tiny-scale (see caveats in
  text).}
\label{tab:terramind_ps4_compare}

\resizebox{\columnwidth}{!}{%
\begin{tabular}{lcccc}
\toprule
\textbf{Dataset}
  & \textbf{TM-ps4 (retrofit)}
  & \textbf{TM-ps16}
  & \textbf{THOR-ps4}
  & \textbf{UNet} \\
\midrule
Sen1Floods11
  & \textbf{0.919} & 0.910 & 0.904 & 0.902 \\
HLS Burn Scars
  & 0.894 & 0.891 & 0.865 & \textbf{0.902} \\
CocoaMining
  & 0.830 & 0.817 & \textbf{0.821} & 0.797 \\
\bottomrule
\end{tabular}%
}

\end{table}

\textbf{Interpretation.} With its best decoder and freeze regime, the
partially re-pretrained TerraMind-ps4 matches or exceeds both
TerraMind-ps16 and THOR-ps4 on Sen1Floods11 (0.919 vs.\ 0.910 and
0.904) and HLS Burn Scars (0.894 vs.\ 0.891 and 0.865), and is
competitive with THOR-ps4 on CocoaMining (0.830 vs.\ 0.821), despite
being a Tiny-scale model compared against Base-scale references and
despite only 48 epochs of partial embedding retraining. This suggests
that at least part of THOR's advantage from finer tokenisation is not
an architectural property exclusive to THOR, but is accessible to
TerraMind's encoder once its patch embedding is adapted to a smaller
patch size -- even without retraining the tokenizer or repeating
TerraMind's full multi-modal pretraining regime.
UNet remains the strongest model on HLS Burn Scars in
absolute terms (0.902), consistent with the pattern discussed in
Section~\ref{sec:low_data} that GFM pretraining confers less advantage
on this particular optical, single-sensor task.

We emphasize that this comparison involves several uncontrolled
differences -- model scale (Tiny vs.\ Base), an incomplete pretraining
regime, and the absence of a validation split during retraining -- and
should be read as a preliminary signal rather than a controlled
ablation. It nonetheless suggests that patch size and pretraining
regime are, to some extent, separable design axes: THOR's fine-patch
advantage and TerraMind's pretraining advantage need not be mutually
exclusive properties of the two architectures. 

\subsection{Cross-Task Synthesis}
\label{sec:synthesis}

Drawing together the preceding analyses, the following picture emerges:

\begin{enumerate}
  \item \textbf{Within the explored configuration space, design choices
    dominate model identity.} Patch size and decoder type jointly
    explain more performance variance than the choice of foundation
    model, on the swept datasets and under the present training recipe
    (Figure~\ref{fig:variance}).

  \item \textbf{Two complementary investment strategies.} TerraMind
    invests at \emph{pretraining time} -- large corpus, many modalities,
    dual-scale loss -- to produce strong fixed-resolution features. THOR
    invests at \emph{inference time} through finer tokenisation. At
    matched compute (ps16), TerraMind consistently leads; THOR closes or
    surpasses the gap only at $4$--$16\times$ higher cost.

  \item \textbf{Decoder importance is resolution-dependent.} At fine
    THOR patch sizes, a linear decoder suffices; at coarse patch sizes
    or for TerraMind, UperNet is essential
    (Figure~\ref{fig:decoder_freeze}).

  \item \textbf{Simple fusion schemes did not reliably outperform
    S2-only on the tested datasets, in the full-data regime.} Shallow
    fusion mechanisms (concatenation, mean pooling) do not reliably
    extract SAR complementarity when optical signal is abundant
    (Figure~\ref{fig:modality}). This may not generalise to
    data-scarce settings.


  \item \textbf{Tiny models offer strong cost--accuracy trade-offs.}
    TerraMind-Tiny + UperNet achieves $>$99\% of Base mIoU at a
    fraction of the cost on Sen1Floods11 (Figure~\ref{fig:scale}).

  \item \textbf{Minority-class tasks favour fine tokenisation.}
    Spatially compact targets (mining, icebergs) benefit
    disproportionately from smaller patch sizes
    (Figure~\ref{fig:s1_anomaly}).
\end{enumerate}

These structural observations -- the trade-off between pretraining
investment and inference-time flexibility, the interaction between
decoder capacity and feature resolution, and the limits of shallow
multimodal fusion in the full-data regime -- suggest hypotheses likely
relevant to other ViT-based GFMs beyond THOR and TerraMind. However,
generalisation beyond the evaluated models, task collection, and
training recipe should be made cautiously.

\section{Conclusion}

This comparative study, spanning ten EO use cases and more than 800 controlled runs, was designed as a natural experiment: two GFMs with contrasting design philosophies, evaluated under a shared protocol, used to ask not \textit{which model wins but what drives the difference}. The main finding is that, within the explored configuration space, \textit{architectural and deployment choices explain more performance variance than model identity itself}: patch size and decoder type jointly account for more variance than the choice of foundation model (Figure~\ref{fig:variance}), and the two interact -- a lightweight decoder suffices once tokenisation is fine, while a multi-scale decoder becomes essential at coarse patch sizes or at a fixed operating resolution. For a practitioner, fixing the compute budget and tokenisation/decoder configuration is at least as consequential as choosing between GFMs, a conclusion we expect to generalise to other ViT-based GFMs.
THOR and TerraMind illustrate two complementary ways of allocating this investment: paying at pretraining time -- a large, multimodal corpus and a dual-scale objective, as in TerraMind -- to obtain strong features at a fixed, moderate resolution; or paying at inference time -- variable-resolution tokenisation, as in THOR -- to trade compute for spatial detail on demand. At matched compute the pretraining-heavy strategy tends to lead, while the inference-adaptive strategy closes or reverses the gap only at substantially higher cost, more reliably so on tasks with spatially compact targets or SAR-dominant signal. Neither is categorically better: which one pays off depends on the binding constraint -- training budget, inference latency, target class geometry, or sensor modality -- and we see this as the more transferable lesson than a ranking between two specific models. Compute-adaptive architectures also carry a structural benchmarking asymmetry worth flagging: their extra tokenisation axis extends the achievable Pareto frontier relative to a fixed-resolution model, and headline comparisons should control for this rather than treat it as incidental.
A few more specific results point to hypotheses we expect to outlast this particular model pair: coarse, fixed tokenisation appears to systematically under-serve spatially compact or minority classes, independent of pretraining quality; shallow fusion (concatenation, mean pooling) did not reliably extract SAR complementarity once optical signal was abundant, motivating cross-attention or gated fusion as a concrete direction; and mapping an unseen sensor onto a pretrained input head remained competitive despite a genuine cross-sensor shift, suggesting spectral compatibility with an existing input head can matter more than native resolution match.
More fundamentally, this study argues for a shift in how GFMs are benchmarked. Inspired by \citet{corley2026knowsstateartgeospatial}, we show that the field lacks the shared protocols and attribution controls needed to compare GFMs at all by performing a controlled, attribution-focused comparison of two contrasting GFMs. Aggregate leaderboards answer a ranking question; they cannot attribute an observed gap to backbone, decoder, tokenisation, or dataset artefact, and several of our findings only become interpretable once each dataset's acquisition protocol, class geometry, and spectral provenance are examined alongside the ablation. Rigorous GFM benchmarking should treat both the systematic ablation of architectural choices and the characterisation of the datasets they are evaluated on as first-class components of evaluation, not preprocessing detail. We hope this offers a template other GFM comparisons can reuse: not a single-number verdict, but a diagnostic methodology and a set of falsifiable hypotheses about what, architecturally, actually moves the needle.\\
\subsection*{
Declaration of generative AI and AI-assisted technologies in the manuscript preparation process
}

{
During the preparation of this work, the authors used Claude to assist with polishing the manuscript text and figures. The authors reviewed and edited the output as needed and take full responsibility for the content of the published article.
}

\bibliographystyle{unsrtnat}
\bibliography{references}
\clearpage
\appendix

\section{Use-cases details}
\label{app:usecases}

\subsection{FAST-EO Use Cases}
\label{sec:fasteo_uc}

\textbf{Sen1Floods11}~\cite{bonafilia2020sen1floods11} is a global flood
mapping benchmark comprising 446 hand-labelled $512 \times 512$~px tiles
at 10~m GSD, built from co-registered Sentinel-1 (SAR) and Sentinel-2
acquisitions covering 11~flood events across six continents. The task is
binary pixel-level segmentation (flooded / not flooded), evaluated with
macro-averaged mIoU. Severe class imbalance makes positive-class IoU the
more sensitive diagnostic. The co-availability of SAR and optical
observations makes it well-suited for multimodal fusion studies.
The dataset provides Sentinel-2 imagery in Level-1C format (13 bands, including B1 coastal aerosol, B9 water vapour, and B10 cirrus). TerraMind can ingest all 13 bands natively; THOR's pretraining used Level-2A products, so the three L1C-only bands are dropped when following the THOR band-mapping protocol, leaving 10 bands.

\textbf{HLS Burn Scars}~\cite{jakubik2023prithvi} provides Harmonised
Landsat--Sentinel-2 (HLS) imagery at 30~m GSD for burn scar delineation
across the contiguous United States. Each $512 \times 512$~px tile
contains six spectral bands; the task is binary segmentation evaluated
with mIoU. As a single-modality dataset, HLS Burn Scars serves as a
reference for purely spectral segmentation, and its coarser 30~m GSD
provides a distinct resolution regime.

\textbf{CocoaMining}~\cite{10982207} targets artisanal gold mining and
cocoa plantation mapping in southern Ghana using fused Sentinel-1,
Sentinel-2, and digital elevation model (DEM) data at 10~m GSD. Each $128 \times 128$~px tile is segmented at the pixel level into three classes (cocoa
plantation, artisanal mine, other), evaluated with macro-averaged mIoU.
The 2022 split is preferred over 2016 because it contains a higher
proportion of artisanal mining pixels ($\sim$7\% vs.\ $\sim$5\% in
2016), providing more positive-class signal during training. Mining sites
nonetheless remain spatially compact relative to the image
extent, where individual footprints often span only a handful pixels of a full tile. This makes fine-grained discrimination a challenging test of each
model's ability to resolve spatially under-represented patterns, and
motivating the patch-size analysis of Section~\ref{sec:anomalies}.


\textbf{Methane Leaks}~
The Methane Benchmark Dataset (MBD) is built by KPLabs from 27
AVIRIS-NG airborne overflights, each providing a 432-band hyperspectral
cube at 2.8~m GSD co-registered with a binary ground-truth (GT) mask
generated by the mag1c methane-enhancement tool. We use the
Sentinel-2-matched variant of MBD (\textbf{MBD-S2}): AVIRIS-NG spectra
are downsampled to 12 bands centred on Sentinel-2 wavelengths and
resampled from 2.8~m to 10~m GSD via Py6S atmospheric
correction, yielding inputs spectrally compatible
with THOR and TerraMind pretraining. A co-registration step
corrects systematic misalignment between the CH$_4$ maps and the
hyperspectral imagery using SIFT feature matching; coordinate-based
registration alone proved insufficient given plumes as small as 50
pixels.
The task is patch-level binary
classification -- methane present (super-emitter + plume) vs.\ background -- evaluated by Overall Accuracy (OA). The pipeline
consists of three steps: (i)~5-fold cross-validation split at the flight
level; (ii)~feature extraction via the frozen encoder followed by global
averaging and a lightweight classification head; and
(iii)~single-class probability output via a fully-connected
head with dropout and softmax activation.

\textbf{Soil Properties -- HYPERVIEW}~\cite{nalepa2023hyperview}
targets the estimation of four soil chemical
properties -- potassium~(K), magnesium~(Mg), phosphorus~(P$_2$O$_5$),
and pH -- from airborne hyperspectral imagery over agricultural fields in
Poland. The task is multi-output regression evaluated per property with
RMSE. With hundreds of narrow spectral bands, HYPERVIEW tests model
generalisation to input modalities that differ substantially in spectral
configuration from the broadband satellite sensors used during
pretraining of both THOR and TerraMind. Results for both Methane Leaks
and HYPERVIEW are reported separately in Section~\ref{sec:hyperspectral},
as both use cases require dedicated discussion of their hyperspectral
input characteristics.

\subsection{FM4CS Use Cases}
\label{sec:fm4cs_uc}

\textbf{Iceberg Detection} focuses on the detection of small maritime
targets in Sentinel-1 SAR imagery, motivated by applications in maritime
safety and climate monitoring. The dataset is derived from the
International Ice Patrol iceberg sightings database, comprising 1355
Sentinel-1 IW image crops ($256 \times 256$~px) with 2191 verified
iceberg annotations collected between 2019 and 2021 over the Labrador
Sea. The task is formulated as binary semantic segmentation (iceberg
vs.\ background) and is evaluated using an instancewise F1 score.
The challenge for this use case lies in detecting small icebergs ($<$60\,m) in
heterogeneous environments with sea ice clutter and speckle noise.

\textbf{Snow Monitoring} focuses on the estimation of key snow
properties for hydrological modelling in high-latitude and mountainous
regions. The task is formulated as pixel-wise regression of
fractional snow cover (FSC). Input consists of multi-band SLSTR imagery
at ${\sim}480$~m GSD, processed as patches of $144 \times 144$~px.
Training labels are derived from high-resolution Sentinel-2-based snow
products (2016--2021), downsampled to SLSTR resolution.

\textbf{Mires/Wetlands Mapping} targets land cover classification with a focus
on wetland detection for environmental monitoring and spatial planning.
The dataset uses Sentinel-2 MSI imagery at 10~m GSD, processed
as $288\times288$ tiles, with sliding-window inference applied during
evaluation. The task is formulated as multi-class semantic segmentation,
with class imbalance handled via weighted loss, and evaluated using mIoU.

\textbf{Flood Zone Mapping} addresses high-resolution flood delineation
using Sentinel-1 SAR imagery at 10~m GSD. The dataset comprises 179
manually annotated flood objects from three major flood events in
southern Norway (2015--2018), with each sample constructed as a pair of
co-registered pre-event and event SAR images. The task is formulated as
a binary segmentation based change detection task.

\textbf{Sea Ice Concentration Estimation} builds on the AI4Arctic Sea
Ice Challenge dataset, targeting semantic segmentation of sea ice for
climate monitoring and maritime safety. In our setting, the task is
simplified to three classes by grouping concentration intervals into
0--10\% (open water), 10--90\% (intermediate ice), and 100\% (packed
ice). The dataset consists of Sentinel-1 SAR observations
over the Danish Meteorological Institute's Greenland monitoring region.

\section{Experimental Setup Details}

\subsection{Loss Functions}
\label{sec:losses}

Loss functions are selected per task type. For segmentation, we use
Dice loss on Sen1Floods11 and HLS Burn Scars, and standard
cross-entropy (without class weighting) on CocoaMining -- the latter
yielding the best empirical results despite the class imbalance in that
dataset.
For the FM4CS use cases, loss functions are likewise matched to task
type: Snow Monitoring, formulated as pixel-wise regression of
fractional snow cover, uses mean absolute error (MAE). Iceberg
Detection and Flood Zone change detection, both binary segmentation
tasks, use standard cross-entropy. Sea Ice Concentration Estimation,
a three-class segmentation task, likewise uses cross-entropy. Mires/Wetlands
Mapping uses a combined Dice~+~cross-entropy loss, which we found
better balances the six-class imbalance handled via weighted loss
described in Section~\ref{sec:fm4cs_uc}.

\subsection{Dataset-Specific Configuration}
\label{sec:dataset_config}

For CocoaMining, we use the 2022 acquisition rather than the 2016
imagery. Neither THOR nor TerraMind incorporates temporal modelling; the
benchmark therefore focuses on single-date semantic segmentation. Mining footprints remain
spatially compact -- typically spanning only a few contiguous pixels at
$128\times128$ resolution -- making this dataset a strong test of fine-grained
discrimination under class imbalance.



To ensure consistent optimization across datasets and model configurations, we used an effective batch size of 8 across all experiments. TerraMind always fits the full effective batch size on a single GPU ($b_\text{phys}=8$) and therefore never requires gradient accumulation. For THOR, smaller patch sizes increase VRAM consumption quadratically, requiring a reduced physical batch size compensated by gradient accumulation via $\texttt{accumulate\_grad\_batches} = 8/b_\text{phys}$. This  affects patch size 4 on HLS Burn Scars and Sen1Floods11, where the physical batch size is reduced to 2, requiring 4 accumulation steps; all other patch sizes (8, 16, 32) on these datasets, as well as all other datasets, retain the standard physical batch size of 8 with no accumulation needed. This reduced-batch schedule at patch size 4 applies uniformly across crop sizes: the $288^2$ crops used for HLS Burn Scars and for frozen Sen1Floods11, and the $224^2$ crops used for unfrozen Sen1Floods11.

\subsection{Low-Data Subset Construction}
\label{sec:subset_construction}

To evaluate performance in data-scarce conditions, we construct 5\%,
10\%, 25\% and 50\% training subsets for Sen1Floods11 and HLS Burn Scars.
The original author-defined validation and test splits are preserved
intact in both cases. Exact training-set sizes per fraction are:
\textbf{Sen1Floods11}: 13 / 26 / 63 / 126 images;
\textbf{HLS Burn Scars}: 25 / 51 / 130 / 262 images.

\textbf{Sen1Floods11.} Random stratified sampling by country tag,
ensuring proportional geographic coverage at the reduced fraction.

\textbf{HLS Burn Scars.} Random subsampling stratified by the
distribution of positive-class pixel ratios across tiles, following
the PANGAEA protocol~\cite{marsocci2024pangaea}.

\textbf{CocoaMining.} Excluded from the low-data ablation: CocoaMining does
not mention a random split in the original repo. HLS Burn Scars 
mentions that tiles are not split spatially, which allowed for a stratified random split
without changing the original protocol. 

Validation is performed every 5 epochs to reduce wall-clock time;
early stopping and LR scheduling patience are set to 10 and 5
validation events respectively. All subset results are averaged over
3 random seeds (seeds 0, 1, 2). These seeds are used when creating splits, 
not at compute time.

\section{TerraMind's Thinking-in-Modalities}
\label{sec:tim_results}

TerraMind's Thinking-in-Modalities mechanism generates synthetic
modality tokens at inference time, either reconstructing a missing
sensor or producing a task-relevant auxiliary modality. We evaluate TiM
on Sen1Floods11 -- the only FAST-EO dataset whose Sentinel-2 band
complement matches TerraMind's pretraining configuration -- and on the
FM4CS use cases, where a LULC TiM modality is evaluated across Flood
Zone Mapping, Iceberg Detection, and Sea Ice Mapping.

\begin{figure*}[t]
\centering
\includegraphics[width=\textwidth]{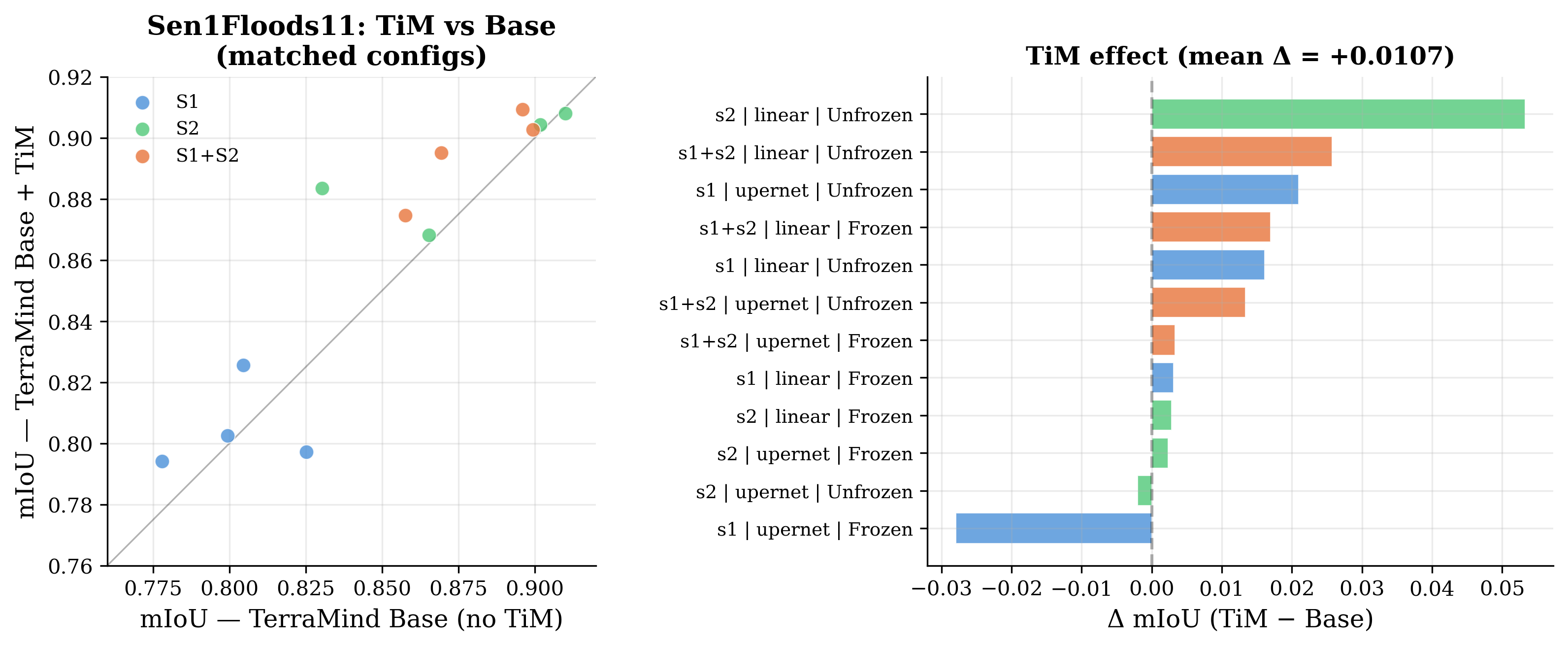}
\caption{%
  \textbf{Effect of Thinking-in-Modalities (TiM) on TerraMind Base
  performance, Sen1Floods11.}
  \textbf{a,} Scatter of mIoU without TiM (x-axis) versus with TiM
  (y-axis) for matched (modality, decoder, freeze-state) configurations.
  \textbf{b,} Per-configuration difference in mIoU (TiM~$-$~base),
  sorted by magnitude.
}
\label{fig:tim}
\end{figure*}

One of the most interesting results on Sen1Floods11 is the
\textbf{decoder-dependent value of TiM}. With UperNet (FT), TiM
S1+S2$\to$LULC achieves 0.909 -- within 0.1~pp of the standard S2-only
optimum (0.910). TiM neither helps nor hurts when a strong multi-scale
decoder is already available. With a \emph{linear decoder}, however,
TiM S1+S2$\to$LULC reaches \textbf{0.895} -- a 6.5~pp improvement over
standard S2 Linear FT (0.830) and 2.6~pp over S1+S2 Linear FT (0.869).

This finding should be understood as an \emph{extension} of the original
TerraMind paper~\cite{jakubik2025terramind}, not a correction of it.
Jakubik et al.\ evaluate TiM under the PANGAEA protocol (frozen backbone
+ UperNet decoder) and report gains of up to 2~pp mIoU. Our result is
fully consistent with this: under the same UperNet regime, TiM provides
a negligible gain ($<$0.1~pp). The larger effect we observe (+6.5~pp)
is specific to the linear decoder condition, a configuration not
investigated in the TerraMind paper. On Sen1Floods11, then, the
governing factor appears to be \textbf{decoder-conditioned utility}:
TiM's synthetic LULC tokens inject semantically structured spatial
information into the token sequence, \emph{partially compensating for
the absent decoder hierarchy} -- a ``soft decoder substitute'' useful
precisely when the downstream architecture is constrained, and
redundant when it is not.

\FloatBarrier

\textbf{The FM4CS use cases show that decoder capacity is not the only,
or even the dominant, modulator of TiM utility.} On \textbf{Sea Ice
Mapping}, TiM improves performance almost irrespective of decoder or
freeze regime: frozen UperNet gains 22.9~pp at Tiny scale (0.526 to
0.755) and 11.0~pp at Base scale (0.558 to 0.668), while frozen Linear
also improves at both scales.

\FloatBarrier

\section{Robustness Diagnostics}
\label{app:robustness}

To assess the robustness of the main-text findings, we report four complementary diagnostics: cross-dataset rank consistency, compute-normalised efficiency, marginal factor contributions, and extended scaling behaviour.

\begin{figure*}[!htb]
\centering
\includegraphics[width=\textwidth]{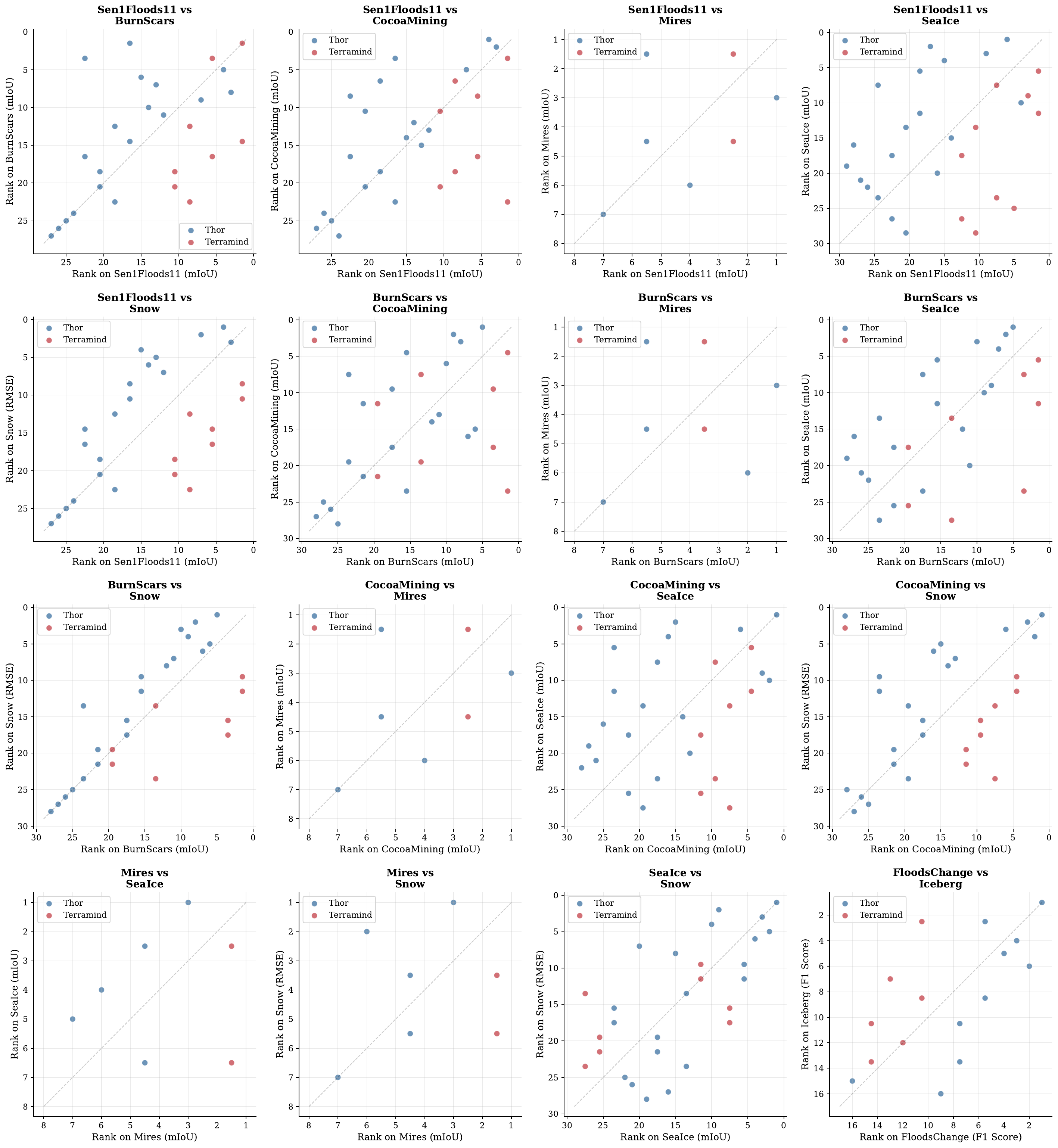}
\caption{%
  \textbf{Cross-dataset rank consistency of configuration performance.}
  Each panel is a scatter plot of per-configuration rank on one dataset
  (x-axis) against rank on another (y-axis), for one within-group
  dataset pair; configurations are matched by backbone size,
  freeze-state, and patch size, independent of decoder (best of the two
  decoders per group is used) and of input modality. Rank~1 denotes the
  best-performing configuration (highest mIoU/F1 score, or lowest RMSE).
}
\label{fig:rank_consistency}
\end{figure*}

\FloatBarrier

\begin{figure*}[!htb]
\centering
\includegraphics[width=\textwidth]{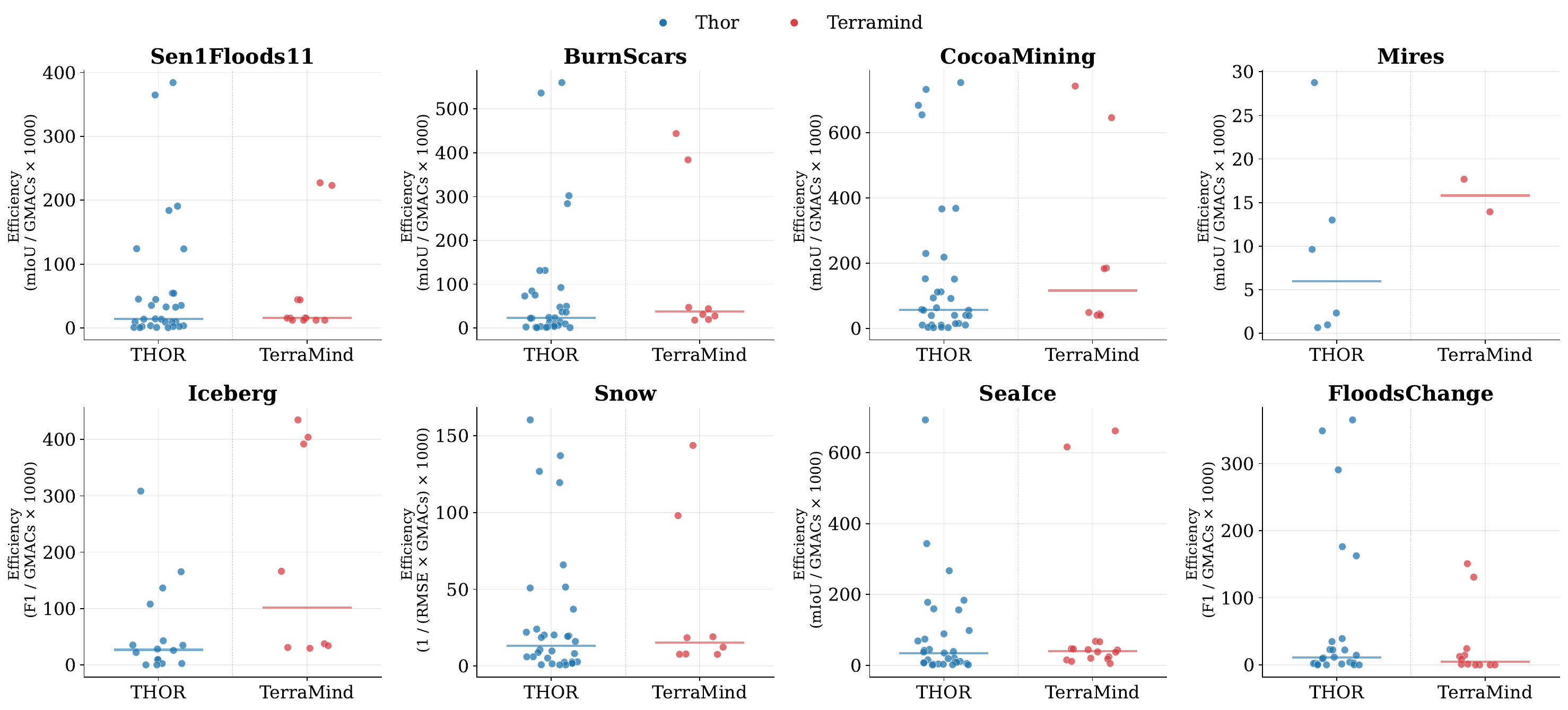}
\caption{%
  \textbf{Compute-normalized efficiency across configurations, by
  dataset.}
  Each dot represents one (model, backbone size, decoder, freeze-state,
  patch size) configuration; horizontal bars mark the group median.
  Efficiency is metric per total (encoder~+~decoder) GMACs, scaled by
  $1000$: mIoU~/~GMACs for mIoU datasets, F1 score~/~GMACs for
  instance-wise F1 datasets, and $1/(\text{RMSE}\times\text{GMACs})$
  for Snow, such that higher values indicate greater efficiency in all
  panels.
}
\label{fig:compute_ranking}
\end{figure*}

Beyond ranking stability, Figure~\ref{fig:rank_consistency} above, we next examine whether these configurations remain efficient once compute cost is taken into account, as shown in Figure~\ref{fig:compute_ranking}.

\FloatBarrier

\begin{figure*}[!htb]
\centering
\includegraphics[width=\textwidth]{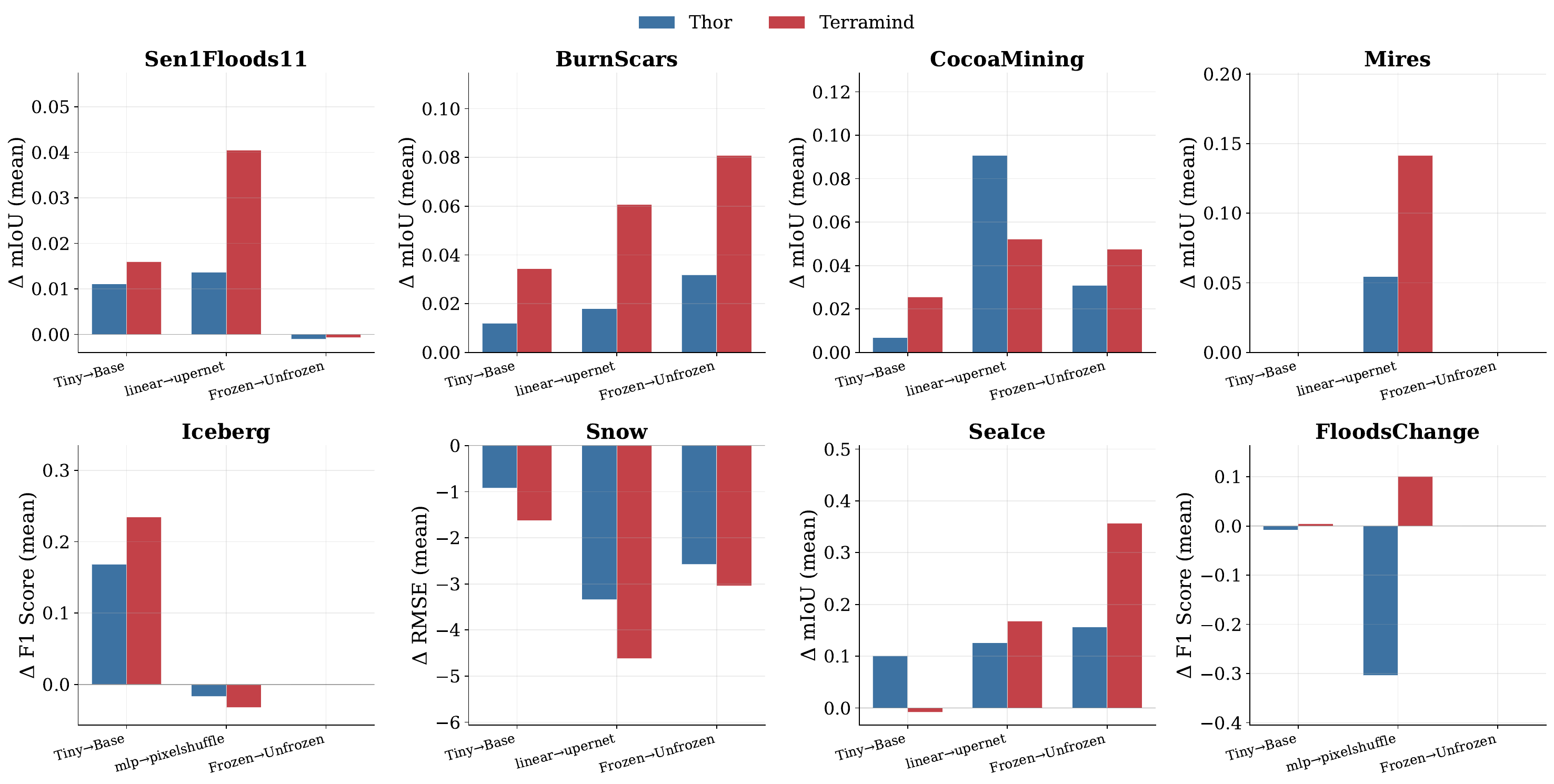}
\caption{%
  \textbf{Marginal value of design upgrades on downstream performance,
  by model and dataset.}
  Bars show the mean change in performance metric when a single design
  factor is switched from a lower to an upper level, averaged over all
  other factors: backbone size (Tiny$\to$Base), decoder (linear$\to$UperNet
  for mIoU/RMSE datasets, MLP$\to$PixelShuffle for instance-wise F1
  datasets), and freeze-state (Frozen$\to$Unfrozen).
}
\label{fig:marginal_value}
\end{figure*}

To disentangle which design choices drive the efficiency differences seen in Figure~\ref{fig:compute_ranking}, we decompose performance into the marginal contribution of each shared upgrade in Figure~\ref{fig:marginal_value}.

\FloatBarrier

\begin{figure*}[!htb]
\centering
\includegraphics[width=\textwidth]{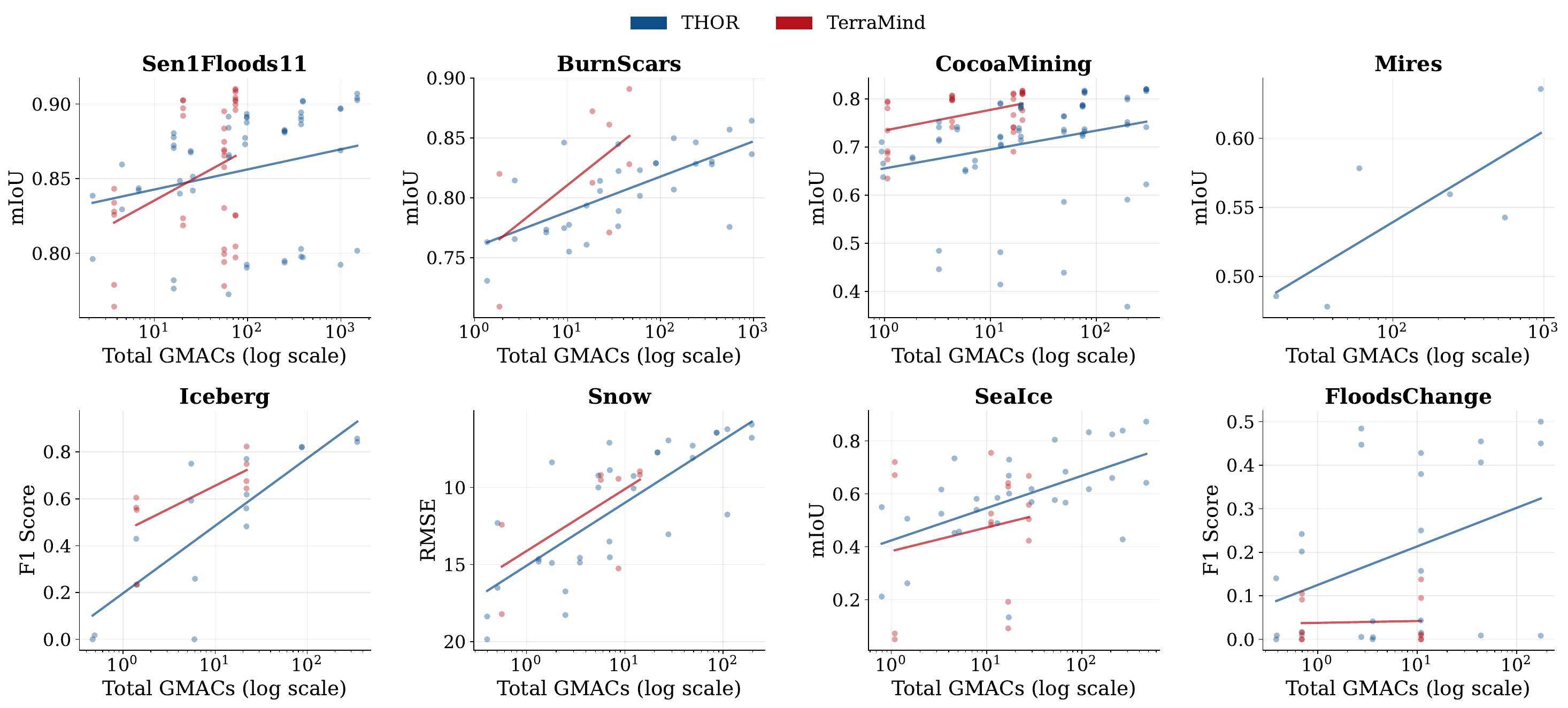}
\caption{%
  \textbf{Computational scaling laws: performance versus total compute,
  by model and dataset.}
  Each panel shows a log-linear fit, metric~$=\alpha\ln(\text{GMACs})+\beta$,
  per model (curve) with underlying configurations (points, all backbone
  sizes, decoders, and patch sizes pooled); $x$-axis is total
  (encoder~+~decoder) GMACs on a log scale. Colors denote model (see
  legend).
}
\label{fig:scaling_extended}
\end{figure*}

Finally, we extend the scaling analysis summarised in Figure~\ref{fig:marginal_value} to smaller model variants, testing whether the observed trends hold at reduced capacity (Figure~\ref{fig:scaling_extended}).

\FloatBarrier

\begin{figure*}[!htb]
\centering
\includegraphics[width=\textwidth]{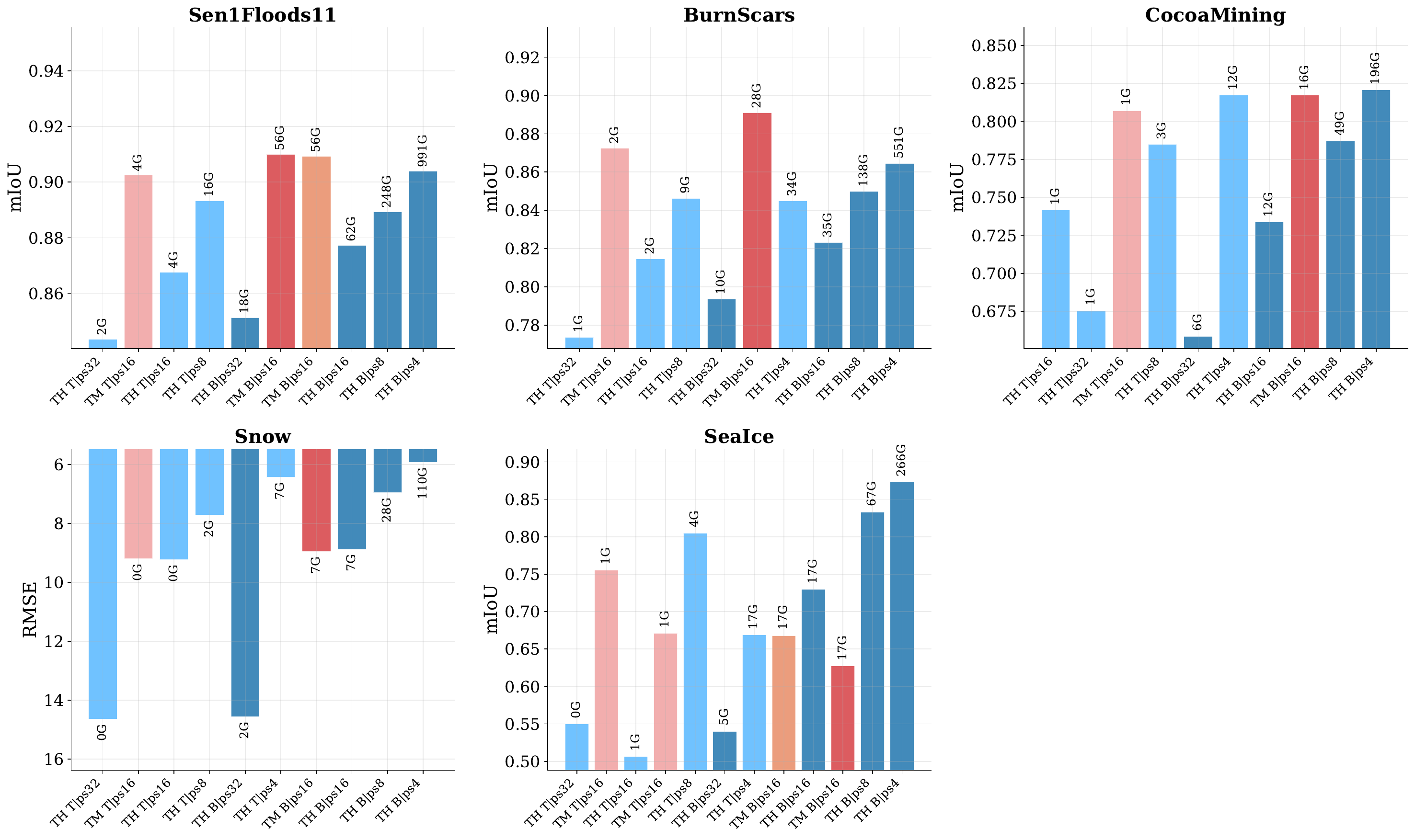}
\caption{%
  \textbf{Performance by configuration, ordered by encoder compute,
  across five datasets.}
  Bars show the best performance metric per (model, backbone size,
  patch size) configuration, for the unfrozen backbone with the better
  of the two decoders per dataset (linear/UperNet for mIoU/RMSE
  datasets, MLP/PixelShuffle omitted here); configurations are sorted
  left to right by increasing encoder GMACs.
}
\label{fig:encoder_efficiency}
\end{figure*}

As a complementary view to the compute-normalised ranking in Figure~\ref{fig:compute_ranking}, we isolate the encoder's own contribution to efficiency, independent of decoder cost, in Figure~\ref{fig:encoder_efficiency}.

\FloatBarrier

\section{GMACs, tokens and patch sizes}
\label{app:graphs}

\begin{figure*}[t]
\centering
\includegraphics[width=\textwidth]{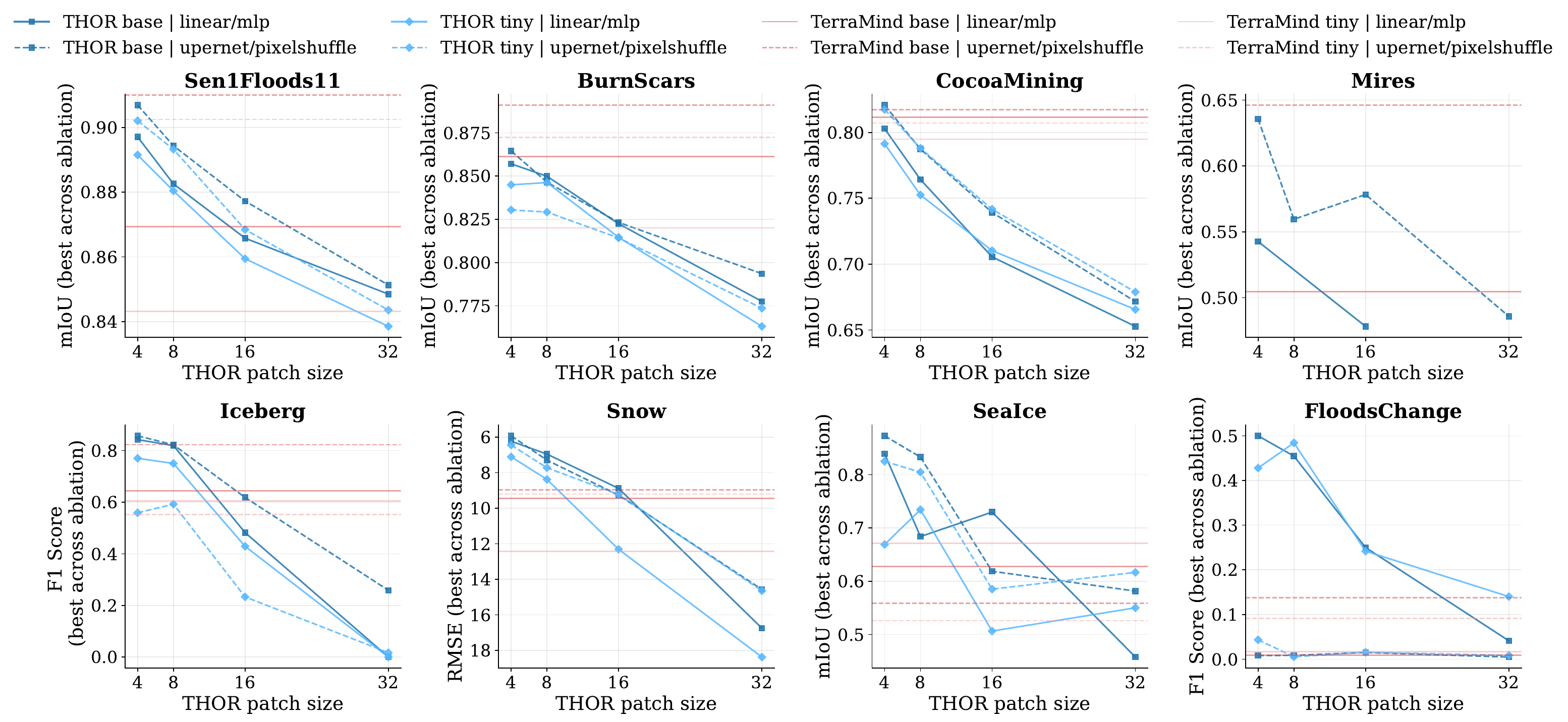}
\caption{%
  \textbf{THOR performance versus patch size, across datasets, with
  TerraMind reference levels.}
  Lines show best performance (across input modality and freeze-state)
  for THOR at each patch size (4, 8, 16, 32), stratified by backbone
  size (Base, Tiny; marker shape) and decoder (solid~=~first-listed
  decoder, dashed~=~second-listed decoder per dataset: linear/UperNet
  for mIoU/RMSE datasets, MLP/PixelShuffle for instance-wise F1
  datasets). Horizontal lines mark best performance for each TerraMind
  configuration (Base, Base+TiM, Tiny, Tiny+TiM; color-coded, labeled at
  right), shown for reference.
}
\label{fig:patchsweep}
\end{figure*}

Figure~\ref{fig:patchsweep} shows the effect of patch size on THOR performance across all FAST-EO and FM4CS datasets, stratified by backbone size and decoder, with TerraMind reference levels included for comparison.

\FloatBarrier

\begin{figure*}[t]
\centering
\includegraphics[width=\textwidth]{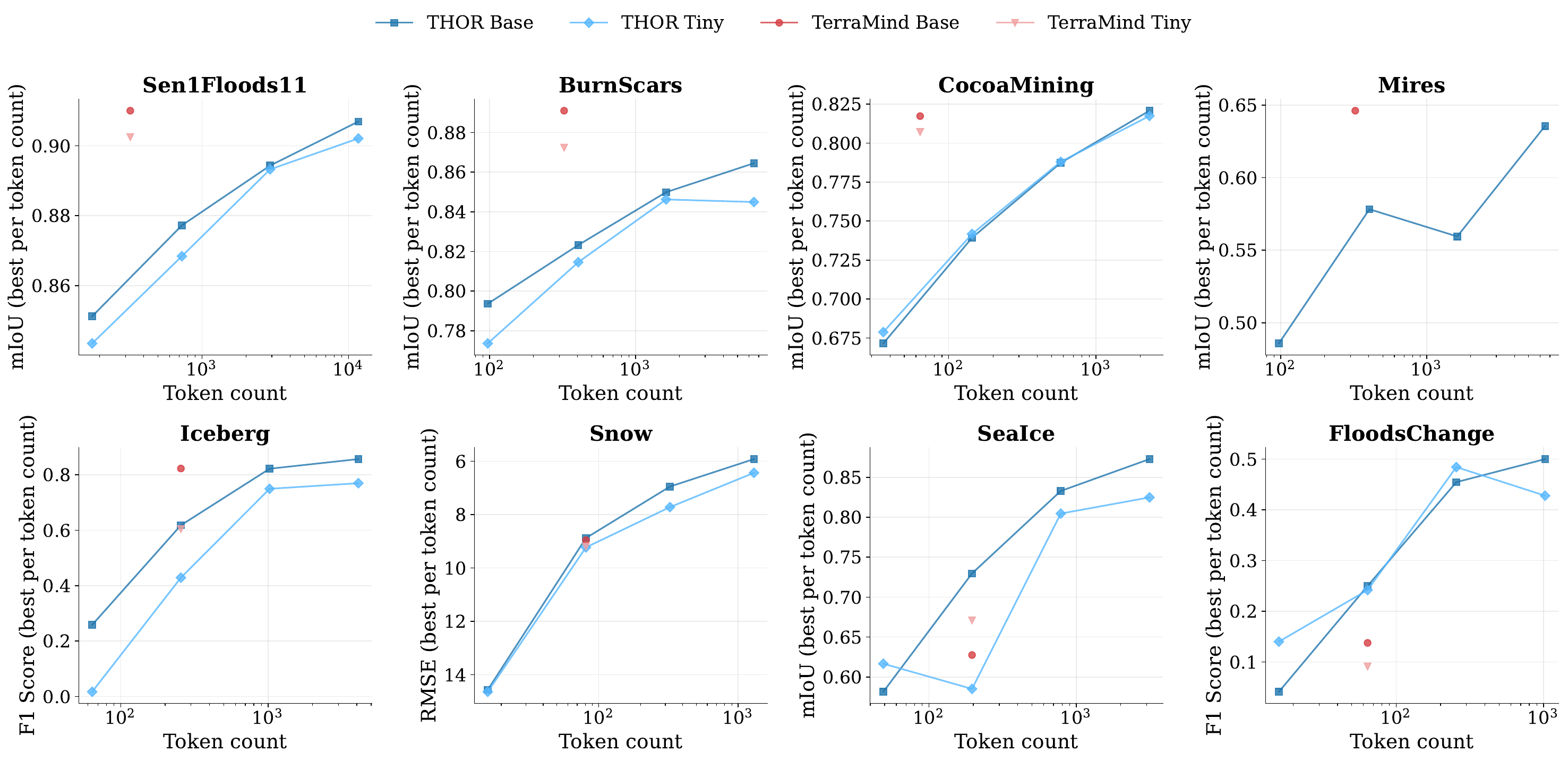}
\caption{%
  \textbf{Performance versus token count, by dataset.}
  Lines show best performance (across all other configuration
  variables) as a function of token count per image (log scale), for
  each model and backbone size (color/marker; see legend), with
  Thinking-in-Modalities (TiM) configurations excluded.
}
\label{fig:scaling}
\end{figure*}

Figure~\ref{fig:scaling} relates downstream performance to the number of tokens per image across datasets and THOR configurations. The figure provides an alternative view of the patch-size experiments, expressing model behaviour in terms of token count rather than spatial patch size.

\FloatBarrier

\section{Token Count vs.\ Spatial Resolution: Controlled Experiments}
\label{app:token_resolution}

Reducing THOR's patch size simultaneously increases the number of tokens \emph{and} the native spatial resolution of each token. Both effects can improve downstream metrics. We designed two experiments to disentangle them.

\textbf{Experiment 1 -- CocoaMining joint resize (not a valid disentanglement).}
We resized train, validation, and test images simultaneously to four target resolutions ($128\times128$, $224\times224$, $288\times288$, $512\times512$) and fine-tuned TerraMind-Base with UperNet. mIoU increases monotonically with resolution (0.796, 0.818, 0.823, 0.835). However, because the \emph{test set is also resized}, the improvement may reflect alignment between the training and evaluation domains rather than richer token representations. Bilinear upsampling (Albumentations \texttt{Resize} uses OpenCV \texttt{INTER\_LINEAR} by default) softens class boundaries, which can inflate mIoU if both train and test are processed identically. \textbf{This experiment does not provide valid evidence that more tokens improve performance. Its main lesson is a benchmark-hygiene warning: resizing evaluation tiles can substantially change reported metrics and must be avoided in fair comparisons.}

\textbf{Experiment 2 -- Sen1Floods11 uptokenized (controlled).}
To avoid the above confound, we take standard $224\times224$ training crops and bilinearly upsample them to $288\times288$, $352\times352$, and $448\times448$ before feeding them to TerraMind-Base, increasing the token count from 196 to 324, 484, and 784 respectively. Using larger crops instead would introduce additional spatial \emph{context} (more scene area per tile), confounding token count with scene coverage; upsampling a fixed-size crop isolates the effect of token count. Validation and test sets remain at the original $224\times224$ resolution with tiled inference. Results are shown in Table~\ref{tab:uptokenized}. Under both frozen and fine-tuned regimes, uptokenized runs degrade slightly relative to the native $224\times224$ baseline. We therefore conclude that naively upsampling training crops does not replicate the benefit of a genuinely finer token grid, and that THOR's ps4/ps8 gains are better attributed to \emph{spatially denser feature extraction at native resolution} than to an increased token budget per se.

\begin{table}[t]
\centering
\caption{Sen1Floods11: effect of upsampling fixed $224\times224$ training crops
to larger resolutions before feeding to TerraMind-Base (UperNet decoder).
Token count = $(H/16)^2$. mIoU on test split ($\uparrow$).
\textbf{Bold}~=~best per regime.}
\label{tab:uptokenized}
\begin{tabular}{lrrr}
\toprule
\textbf{Input fed to model} & \textbf{Tokens} & \textbf{Frozen} & \textbf{FT} \\
\midrule
Native $224\times224$ (no upsampling) & 196 & \textbf{0.905} & \textbf{0.908} \\
Upsampled $224\to288\times288$        & 324 & 0.899 & 0.904 \\
Upsampled $224\to352\times352$        & 484 & 0.895 & 0.905 \\
Upsampled $224\to448\times448$        & 784 & 0.892 & 0.903 \\
\bottomrule
\end{tabular}
\end{table}

\FloatBarrier


\section{Data Efficiency: Extended Analysis}
\label{app:data_efficiency}

\begin{figure*}[t]
\centering
\includegraphics[width=0.88\textwidth]{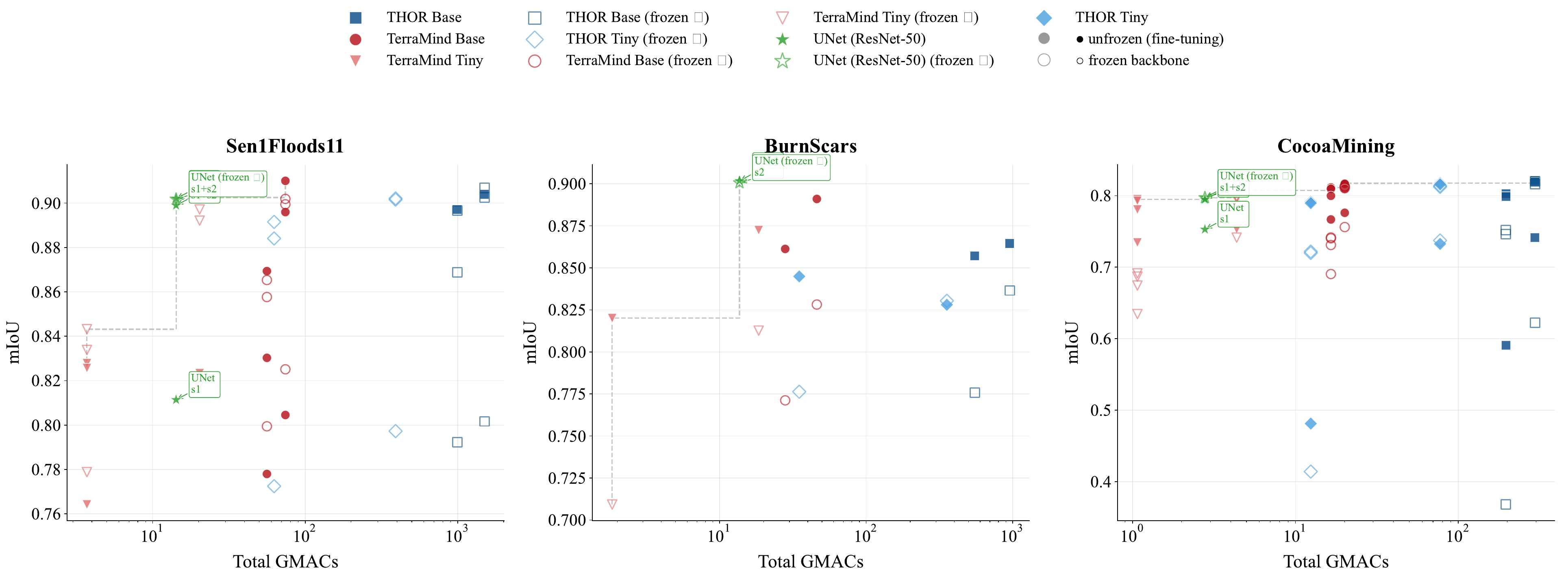}
\caption{%
  \textbf{Performance versus total compute (GMACs), including a UNet
  baseline, across three datasets.}
  Each point is one (model, backbone size, freeze-state) configuration
  at full training data; filled markers denote unfrozen (fine-tuned)
  backbones, open markers denote frozen backbones. UNet is evaluated
  per input modality, with the best-performing modality annotated
  beside its marker. Colors and marker shapes denote model and backbone
  size (see legend). Black dashed step line traces the Pareto frontier
  across all plotted conditions. For Snow (RMSE) and Iceberg
  (instance-wise F1 score), the metric differs from mIoU as indicated
  on the $y$-axis; for Snow, the $y$-axis is inverted so that better
  performance appears toward the top, consistent with other panels.
}
\label{fig:gmacs_pareto}
\end{figure*}

\begin{figure*}[t]
\centering
\includegraphics[width=1\textwidth]{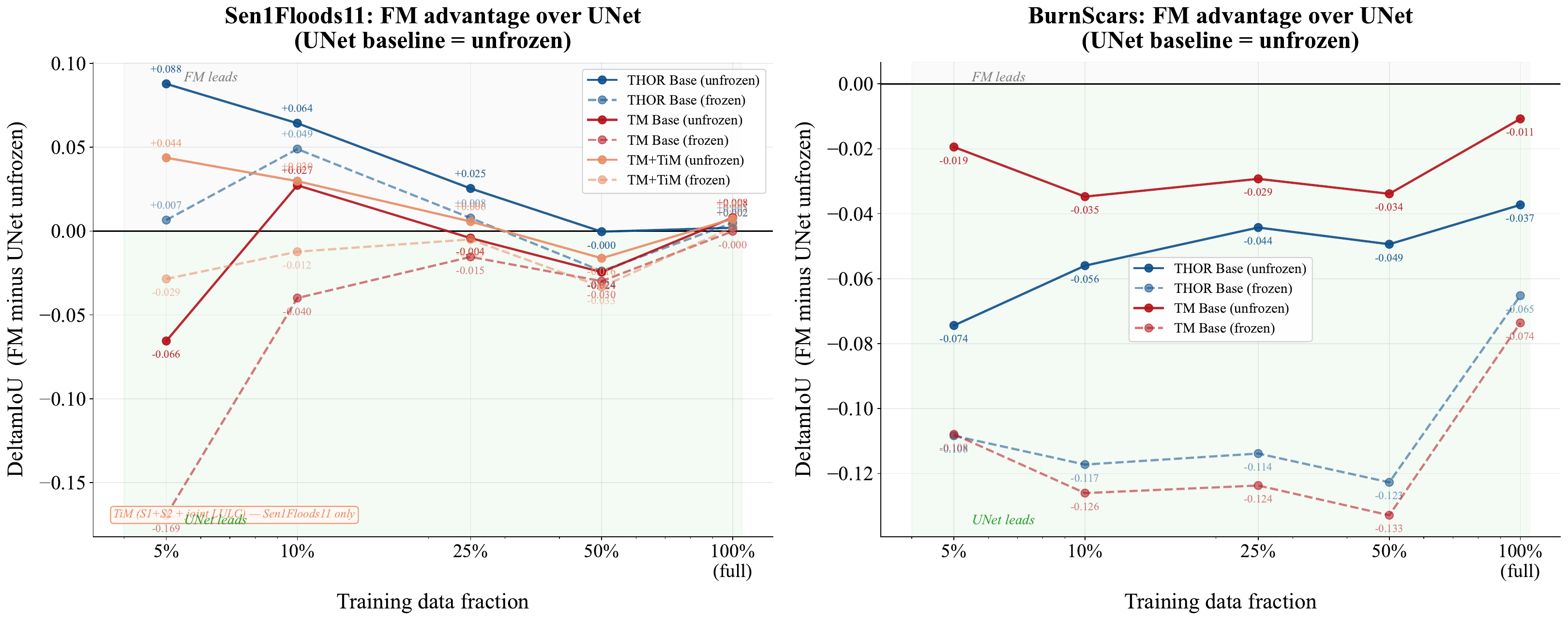}
\caption{%
  \textbf{Performance gap between foundation models and a UNet baseline
  across training data fractions, for Sen1Floods11 (left) and BurnScars
  (right).}
  Lines show $\Delta$mIoU (foundation model minus unfrozen UNet
  baseline) as a function of training data fraction (5, 10, 25, 50,
  100\%); values are annotated at each point. Solid lines denote
  unfrozen (fine-tuned) backbones, dashed lines denote frozen backbones;
  colors denote model (THOR~Base, TerraMind~Base, and, for
  Sen1Floods11, TerraMind+TiM; see legend). Horizontal line at zero
  separates the shaded regions where the foundation model outperforms
  UNet ("FM leads," upper) from where UNet outperforms the foundation
  model ("UNet leads," lower). On Sen1Floods11, TerraMind+TiM is
  evaluated with the S1+S2 modality and joint land-use/land-cover
  (LULC) supervision.
}
\label{fig:gap_vs_unet}
\end{figure*}

Supplementary analyses of computational efficiency and training-data requirements. Figure~\ref{fig:gmacs_pareto} relates performance to total compute, while Figure~\ref{fig:gap_vs_unet} shows how the advantage (or disadvantage) of foundation models relative to a UNet baseline evolves across training-data fractions.

\FloatBarrier

\FloatBarrier

\section{Task-Specific UNet Baseline: Full-Dataset Results}
\label{app:unet_results}

Table~\ref{tab:unet_full} compares a UNet encoder-decoder trained from
scratch against the best GFM configurations on the full training set of
the three investigated FAST-EO datasets. On Sen1Floods11, both GFMs outperform
UNet (+2.8--3.1\,pp), consistent with the multimodal (S1+S2) pretraining
advantage on a multimodal flood-mapping task. On HLS Burn Scars, UNet
(0.901) outperforms both THOR (0.865) and TerraMind (0.891) even at
full data -- the same ordering observed at every fraction in
Section~\ref{sec:low_data} -- which we attribute to the HLS source
distribution shift and the absence of SAR input (see
Section~\ref{sec:low_data}). On CocoaMining, GFMs lead by
3.2--3.5\,pp; the fine-grained spatial structure of artisanal mining
sites (covering a small fraction of each $128\times128$ patch) rewards
the dense feature extraction of GFMs at ps4.

\begin{table}[t]
\centering
\caption{Full-dataset mIoU for the UNet baseline vs.\ best GFM
configuration (UperNet decoder). UNet is trained from scratch.
\textbf{Bold}~=~best per dataset. GFM best values are the maximum
observed across all backbone/freeze/patch-size configurations.}
\label{tab:unet_full}
\begin{tabular}{lrrr}
\toprule
\textbf{Dataset} & \textbf{UNet} & \textbf{Best THOR} & \textbf{Best TM} \\
\midrule
Sen1Floods11   & 0.879 & 0.907 & \textbf{0.910} \\
HLS Burn Scars & \textbf{0.901} & 0.865 & 0.891 \\
CocoaMining    & 0.785 & 0.820 & \textbf{0.817} \\
\bottomrule
\end{tabular}
\end{table}
\FloatBarrier
\section{TerraMind at Patch Size 4 Results}
\label{app:tmps4results_results}

Tables~\ref{tab:full_results} and \ref{tab:full_results_ft} compare the partially re-pretrained TerraMind Tiny PS4 variant against the standard TerraMind and THOR models. Relative to the native TerraMind Tiny model, the PS4 variant improves performance across all three FAST-EO datasets under both frozen-backbone and full fine-tuning settings, and achieves the strongest results on Sen1Floods11 and HLS Burn Scars. 


\begin{table}[t]
\centering
\caption{%
  FAST-EO results (best frozen backbone per model).
  Segmentation: macro-averaged mIoU. 
  The best configuration is selected by sweep over patch size and
  decoder for Thor. 
  \textbf{Bold} = best result per column.}
\label{tab:full_results}
\begin{tabular}{lccc}
\toprule
\textbf{Model}
  & \textbf{Sen1Fl.} & \textbf{Burn Scars} & \textbf{Cocoa} \\
  & mIoU & mIoU & mIoU \\
\midrule
TerraMind Base 
  & 0.904 & 0.828 & 0.813 \\
TerraMind Tiny 
  & 0.897 & 0.813 & 0.799 \\
\textbf{Terramind Tiny PS4} 
  & \textbf{0.916} & \textbf{0.845} & 0.805 \\
\midrule
THOR Base 
  & 0.907 & 0.837 & \textbf{0.820}\\
THOR Tiny 
  & 0.902 & 0.830 & 0.814 \\
\bottomrule
\end{tabular}%

\end{table}

\begin{table}[t]
\centering
\caption{%
  Full benchmark results (best full fine-tuning per model).
  Segmentation: macro-averaged mIoU.
  All results use fully fine-tuned backbone and the best patch size (for Thor) /
  decoder found per dataset.
  \textbf{Bold} = best result per column.}
\label{tab:full_results_ft}
\begin{tabular}{lccc}
\toprule
\textbf{Model}
  & \textbf{Sen1Fl.} & \textbf{Burn Scars} & \textbf{Cocoa}\\
  & mIoU & mIoU & mIoU \\
\midrule
TerraMind Base
  & 0.910 & 0.891 & 0.817\\
TerraMind Tiny
  & 0.902 & 0.872 & 0.807 \\
\textbf{Terramind Tiny PS4}
  & \textbf{0.919} & \textbf{0.895} & \textbf{0.829} \\
\midrule
THOR Base
  & 0.904 & 0.865 & 0.821 \\
THOR Tiny
  & 0.893 & 0.846 & 0.817 \\
\bottomrule

\end{tabular}%

\end{table}


\FloatBarrier

\section{Further Considerations and Design Philosophy}

\subsection{Pretraining Domain Emphasis and Consortium Provenance}
\label{sec:domain_emphasis}

Each model tends to perform comparatively better on the use
cases contributed by its own consortium. THOR's pretraining includes a
dedicated SAR reconstruction objective and a sampling strategy biased
toward Arctic geographies within an otherwise global corpus, and the
FM4CS use cases it contributed are themselves concentrated on SAR
sensing and Arctic or sub-Arctic domains (Iceberg Detection, Flood Zone
Mapping, Sea Ice Mapping). THOR leads TerraMind on all three. Conversely, TerraMind's
pretraining is comparatively broadband-optical and multi-modal but not
SAR- or Arctic-biased in the same targeted way, and it is
competitive with or ahead of THOR on the optical, globally-sampled
FAST-EO use cases it contributed (Sen1Floods11, HLS Burn Scars,
Wetland Mapping), particularly at matched patch size
(Section~\ref{sec:patch_size}).

This correspondence is consistent with the hypothesis that
\emph{targeted pretraining-domain emphasis} -- a dedicated SAR
reconstruction head and Arctic-biased sampling for THOR -- confers a
measurable downstream advantage on tasks that share that domain, and
that this advantage persists even when the encoder is frozen
(Section~\ref{sec:frozen_results}), which would suggest the effect is
carried by the pretrained representations themselves rather than by
task-specific adaptation.

We are, however, unable to attribute the pattern to pretraining
emphasis alone. THOR and its FM4CS use cases were developed
jointly within the same consortium, as were TerraMind and its 
FAST-EO use cases. This co-development introduces confounds that are
difficult to separate from a domain-emphasis effect: use-case selection
may implicitly favour scenarios where the accompanying model is known
to perform well; annotation protocols, sensor pre-processing, and
labelling conventions may be more closely matched to each model's
expected input format within its own consortium; and dataset difficulty
is not calibrated across consortia. The consortium-of-origin variable is
therefore confounded with pretraining-domain emphasis by construction,
and the present design cannot separate the two.

Two observations offer partial, indirect support for a genuine
pretraining effect rather than a purely consortium-driven artefact.
First, the advantage is domain-specific rather than uniform: THOR does
not lead on every FM4CS use case regardless of modality (Wetland
Mapping, evaluated only within the FAST-EO set, is the one FAST-EO task
where the coarser-patch-size THOR configuration is outperformed by
TerraMind, Section~\ref{sec:overview}), and TerraMind's own frozen
representations degrade sharply on SAR-only Sea Ice Mapping relative to
its consistently modest frozen--fine-tuned gap elsewhere
(Section~\ref{sec:frozen_results}) -- a pattern more easily explained by
differential SAR pretraining exposure than by consortium provenance
alone. Second, the Snow Monitoring result (Section~\ref{sec:cross_sensor}),
a use case using a sensor (Sentinel-3) present in THOR's pretraining but
absent from TerraMind's, favours THOR by a comparable margin to the
other FM4CS use cases, consistent with a sensor-coverage explanation.
Neither observation, however, constitutes a controlled test: a
definitive attribution would require evaluating both models on
use cases curated independently of either consortium, which we leave to
future work.

We therefore treat the consortium-correlated performance pattern as a
hypothesis meriting the same diagnostic caution urged throughout this
paper (Section~\ref{sec:anomalies}): it is consistent with, but not
proof of, a causal effect of targeted pretraining-domain emphasis on
downstream transfer, and should not be read as evidence that either
model's pretraining recipe is unconditionally superior.
\subsection{Task-Specific Observations and Anomalies}
\label{sec:anomalies}

The preceding analysis identified broad trends. Several dataset-specific
observations merit dedicated discussion.

\begin{figure*}[t]
\centering
\includegraphics[width=\textwidth]{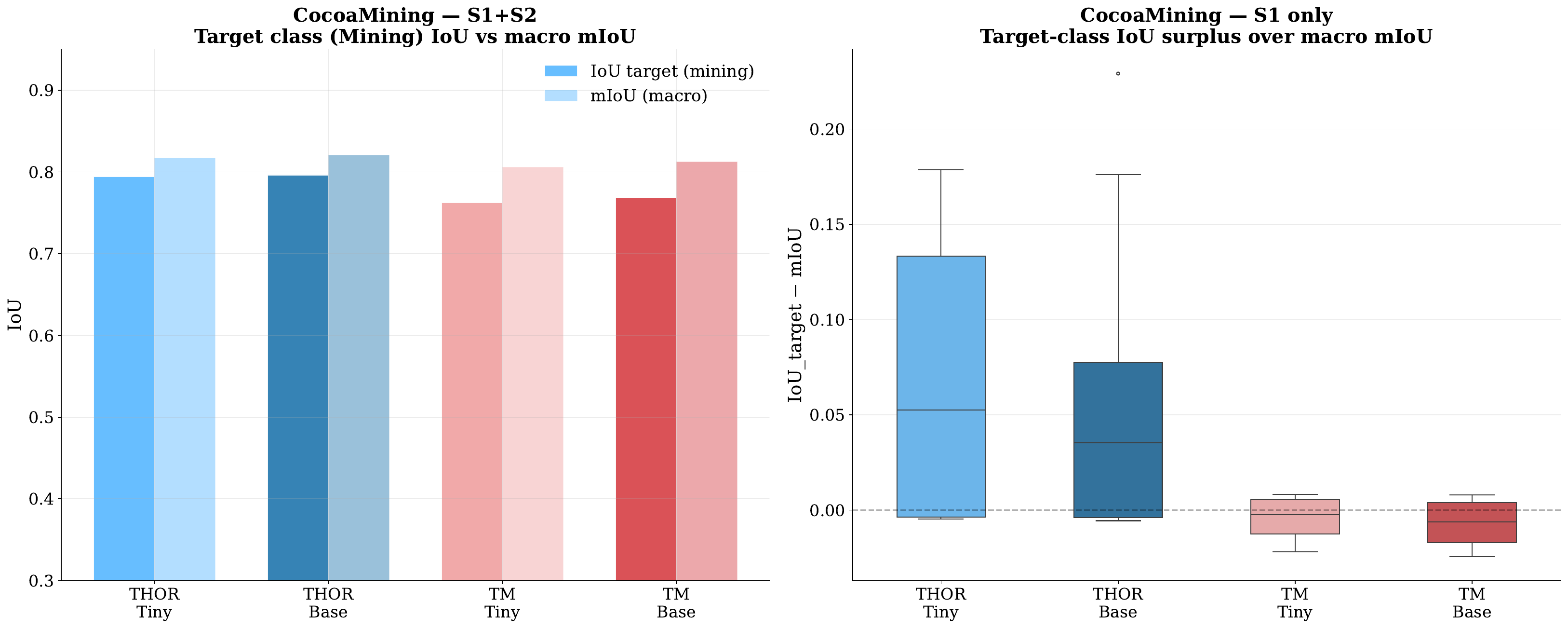}
\caption{%
  \textbf{Target-class (mining) performance relative to macro mIoU on
  CocoaMining.}
  \textbf{a,} Best target-class IoU (dark bars) and macro mIoU (light
  bars) per model and backbone size, for the S1+S2 modality.
  \textbf{b,} Distribution of the gap between target-class IoU and
  macro mIoU (IoU$_{\text{target}}$~$-$~mIoU) across configurations per
  model and backbone size, for the S1-only modality..
}
\label{fig:s1_anomaly}
\end{figure*}

\textbf{CocoaMining: the minority-class--patch-size interaction.}
Figure~\ref{fig:s1_anomaly} illustrates the relationship between patch
size and class-level IoU on CocoaMining.

\FloatBarrier

CocoaMining's artisanal mining class occupies a small fraction of each
$128\times128$ tile. At ps32, each token covers $32 \times 32 = 1{,}024$ native
pixels -- larger than many individual mining sites. The token-level
representation therefore averages over mixed mining/vegetation
boundaries, suppressing the discriminative spectral features. Reducing
to ps4 ($4 \times 4 = 16$ pixels per token) allows the backbone to
represent mining footprints at a resolution commensurate with their
spatial extent. This \emph{minority-class--patch-size interaction} is
likely relevant to other tasks involving spatially compact targets
(small floods, individual icebergs, narrow burn perimeters).

\textbf{THOR outperforms TerraMind on CocoaMining at a smaller native
input.}
A noteworthy result is that THOR at ps4 -- operating at the native
$128\times128$ resolution and setting
\texttt{ground\_covers}~$=1{,}280$\,m to faithfully represent the 10\,m
GSD -- achieves higher mIoU than TerraMind, which must upsample images to
$224\times224$ to meet its minimum pretraining resolution. Upsampling to
match a fixed input size distorts the native spatial structure of the
scene and introduces interpolation artefacts; this is particularly
problematic for a task where distinguishing small artisanal mining sites
from surrounding vegetation relies on locally coherent spectral
signatures. Furthermore, resizing the image before feeding it into
TerraMind effectively increases the reported GMACs for this dataset (the
encoder now processes $196$~tokens instead of $64$), yet accuracy does
not surpass THOR. This result supports the practical value of THOR's
GSD-aware architecture for tasks where preserving native resolution is
physically meaningful.
\textbf{Flood Segmentation: a high-resolution-extraction regime.}
On the FM4CS Flood Segmentation task, both THOR and TerraMind degrade
severely when patch sizes fall below~8, but the pattern does not
resemble a smooth performance gradient: it is a threshold effect,
with a sharp separation between coarse and fine tokenisation regimes.
We interpret this as evidence that flood-zone delineation in this
dataset is driven primarily by fine-grained, locally discriminative
feature extraction rather than by the semantically rich, broad-context
scene understanding that benefits tasks such as Mires/Wetlands Mapping or
Sen1Floods11. Under this interpretation, THOR's compute-adaptive
tokenisation is not merely an incremental improvement but opens a
structurally different approach to this class of problem, one that
TerraMind's fixed ps16 regime cannot access. This result should,
however, be read as a diagnostic rather than a definitive ranking:
because TerraMind at ps16 and THOR at ps4/ps8 operate in
non-comparable resolution regimes, the observed gap does not by
itself establish which architecture is better suited to flood
segmentation in general. It remains open whether an alternative
formulation of this use case -- one that does not depend on
high-resolution local extraction, e.g.\ through coarser labelling
or larger target objects -- would narrow or eliminate the gap under a
coarser ViT patch embedding. We leave this reformulation to future
work.

\textbf{Sea Ice Mapping: class imbalance and out-of-distribution
signalling.} Despite reducing the original 11-class labelling scheme
to 3~classes, Sea Ice Mapping still exhibits pronounced imbalance
between the edge classes (open water, packed ice) and the dominant
transitional class. Because sea ice concentration is spatially and
temporally autocorrelated, constructing a stratified train/val/test
split is non-trivial, and the resulting imbalance manifests as a
counter-signal: validation performance and test performance diverge
across multiple configurations, indicating that the two splits draw
from meaningfully different distributions rather than merely
differing in sampling noise. This divergence can be read either as a
genuine out-of-distribution generalisation gap or as an artefact of
an uncalibrated split; the limited number of runs per configuration
does not allow us to distinguish between these interpretations, and
we therefore treat performance rankings on this dataset with
corresponding caution.

\textbf{Implications for benchmark composition.} We include both Flood
Segmentation and Sea Ice Mapping despite their limited cross-model
comparability, because their diagnostic value lies less in absolute
ranking and more in what they reveal under the intra-model ablation
grid -- patch-size thresholds for THOR, split sensitivity for
TerraMind and THOR alike. This motivates a broader point about
benchmark curation: a holistic evaluation suite should combine simple,
balanced use cases, which support clean interpretation of
representation and adaptation ability, with more nuanced, real-world
use cases that expose the intricacies -- class imbalance, split
instability, resolution dependence -- that practitioners actually
encounter in deployment. The latter category demands more careful
interpretation, and, critically, more explicit communication of its
limitations to the scientific community. We argue that this
communicative dimension is currently underserved by benchmarks
organised primarily around leaderboard-style aggregate metrics, which
reward performance without surfacing the conditions under which that
performance is or is not meaningful.

\subsection{Motivation for Low-Data Evaluation}
\label{sec:low_data_motivation}

The full-data benchmark of the preceding sections evaluates both models
under conditions that may systematically underrepresent THOR's primary
design objective. THOR was explicitly optimised for data-scarce regimes:
Forgaard et al.~\cite{forgaard2026thor} report state-of-the-art results
on the PANGAEA 10\% benchmark and frame compute-adaptive patching as a
direct response to the \emph{data-hungry decoder} problem -- the
observation that standard dense prediction heads require large labelled
datasets to converge, while fine tokenisation provides the decoder with
a denser gradient signal from fewer images.

In operational EO settings, large labelled datasets are the exception
rather than the rule. Annotating flood extents, burn scars, or mining
sites at scale is costly and geographically uneven.
We therefore complement the full-data results with a controlled
low-data ablation, varying the training fraction across several levels.
This allows us to characterise not only which model performs better
under data scarcity, but also \emph{at what data volume} the relative
ranking changes -- a more actionable insight for practitioners deciding
which model to deploy under a given annotation budget.

\end{document}